\documentclass[10pt,twocolumn,letterpaper]{article}

\usepackage[pagenumbers]{cvpr} 

\usepackage{amssymb}
\usepackage{pifont}
\usepackage{microtype}
\usepackage{silence}
\usepackage{listings}
\usepackage{xcolor}

\usepackage{graphicx}
\usepackage{booktabs}
\usepackage{multirow}
\usepackage[table]{xcolor}
\usepackage{subcaption}
\usepackage{amsmath,amssymb}
\usepackage[ruled,vlined]{algorithm2e}
\usepackage{float}
\usepackage{enumitem}
\usepackage{makecell}
\usepackage{array}

\definecolor{lightblue}{RGB}{221,235,247}
\definecolor{overlayorange}{RGB}{255,210,179}
\definecolor{overlayred}{RGB}{255,179,179}
\definecolor{overlaygreen}{RGB}{179,255,179}

\definecolor{niceorange}{HTML}{FBECDD}

\definecolor{ratioThirty}{HTML}{F2F7FB}
\definecolor{ratioFifty}{HTML}{E7F0F8}
\definecolor{ratioHundred}{HTML}{D3E2F2}

\newcommand{\focusui}{\textup{\textsc{FocusUI}}}
\newcommand{\focusuithreebillion}{\textup{\textsc{FocusUI-3B}}}
\newcommand{\focusuisevenbillion}{\textup{\textsc{FocusUI-7B}}}

\newcommand{\focusuitwobillionqwenthreevl}{\textup{\textsc{FocusUI-Qwen3-VL-2B}}}

\newcommand{\pospad}{\textup{\textsc{PosPad}}}
\newcommand{\pospadtoken}{\texttt{<pos\_pad>}}

\newcommand{\tighttt}[1]{\scalebox{0.9}[1]{\texttt{#1}}}

\newcommand{\cmark}{\text{\ding{51}}} 
\newcommand{\xmark}{\text{\ding{55}}} 

\makeatletter
\renewcommand*{\@fnsymbol}[1]{%
  \ensuremath{%
    \ifcase#1
    \or \dagger      
    \or \ddagger     
    \or \mathsection 
    \or \mathparagraph
    \or \|
    \or **
    \or \dagger\dagger
    \or \ddagger\ddagger
    \else\@ctrerr
    \fi
  }%
}
\makeatother

\definecolor{codegreen}{rgb}{0,0.6,0}
\definecolor{codegray}{rgb}{0.5,0.5,0.5}
\definecolor{codepurple}{rgb}{0.58,0,0.82}
\definecolor{backcolour}{rgb}{0.95,0.95,0.92}

\lstdefinestyle{promptstyle}{
    backgroundcolor=\color{backcolour},   
    commentstyle=\color{codegreen},
    keywordstyle=\color{magenta},
    numberstyle=\tiny\color{codegray},
    stringstyle=\color{codepurple},
    basicstyle=\ttfamily\footnotesize,
    breakatwhitespace=false,         
    breaklines=true,
    breakindent=0pt,                 
    captionpos=b,                    
    numbersep=5pt,                  
    showspaces=false,                
    showstringspaces=false,
    showtabs=false,                  
    tabsize=1
}

\definecolor{cvprblue}{rgb}{0.21,0.49,0.74}
\usepackage[pagebackref,breaklinks,colorlinks,allcolors=cvprblue]{hyperref}

\title{\textsc{FocusUI}: Efficient UI Grounding via Position-Preserving Visual Token Selection}

\author{
\textsuperscript{1}Mingyu Ouyang,
\textsuperscript{2}Kevin Qinghong Lin,
\textsuperscript{1}Mike Zheng Shou\thanks{Corresponding authors.}~,
\textsuperscript{1}Hwee Tou Ng\footnotemark[1]
\\
\textsuperscript{1}National University of Singapore\quad
\textsuperscript{2}University of Oxford
\\
{\tt\small ouyangmingyu04@u.nus.edu, \{kevin.qh.lin, mike.zheng.shou\}@gmail.com,  dcsnght@nus.edu.sg}
\\
\href{https://showlab.github.io/FocusUI}{\textcolor[RGB]{104,52,154}{\nolinkurl{https://showlab.github.io/FocusUI}}}
}

\begin{document}

{
\maketitle
}

\vspace{-1em}

\vspace{-0.25em}
\begin{abstract}
    Vision-Language Models (VLMs) have shown remarkable performance in User Interface (UI) grounding tasks, driven by their ability to process increasingly high-resolution screenshots.
    However, screenshots are tokenized into thousands of visual tokens (\eg, about 4700 for 2K resolution), incurring significant computational overhead and diluting attention. 
    In contrast, humans typically \textbf{focus} on regions of interest when interacting with UI.
    In this work, we pioneer the task of efficient UI grounding.
    Guided by practical analysis of the task’s characteristics and challenges, we propose \textup{\textsc{FocusUI}}, an efficient UI grounding framework that selects patches most relevant to the instruction, while preserving positional continuity for precise grounding. 
    \textup{\textsc{FocusUI}} addresses two key challenges:
    \textbf{(1) Eliminating redundant tokens in visual encoding.} 
    We construct patch-level supervision by fusing an instruction-conditioned and a rule-based UI-graph score that down-weights large homogeneous regions to select distinct and instruction-relevant visual tokens.
    \textbf{(2) Preserving positional continuity during visual token selection.}
    We find that general visual token pruning methods suffer from severe accuracy degradation on UI grounding tasks due to breaking positional information.
    We introduce a novel \textup{\textsc{PosPad}} strategy, which compresses each contiguous sequence of dropped visual tokens into a single special marker placed at the sequence’s last index to preserve positional continuity.
    Comprehensive experiments on four grounding benchmarks demonstrate that \textup{\textsc{FocusUI}} surpasses GUI-specific baselines. On the ScreenSpot-Pro benchmark, \textup{\textsc{FocusUI-7B}} achieves performance improvement of 3.7\% over GUI-Actor-7B. Also, even with only \textbf{30\%} visual token retention, the performance of \textup{\textsc{FocusUI-7B}} only drops by 3.2\%, while achieving up to \textbf{1.44$\times$} faster inference and \textbf{17\%} lower peak GPU memory.
\end{abstract}

\vspace{-1em}
    
\begin{figure}[!t]
    \centering
    \begin{subfigure}[t]{\linewidth}
    \centering
    \includegraphics[width=\linewidth]{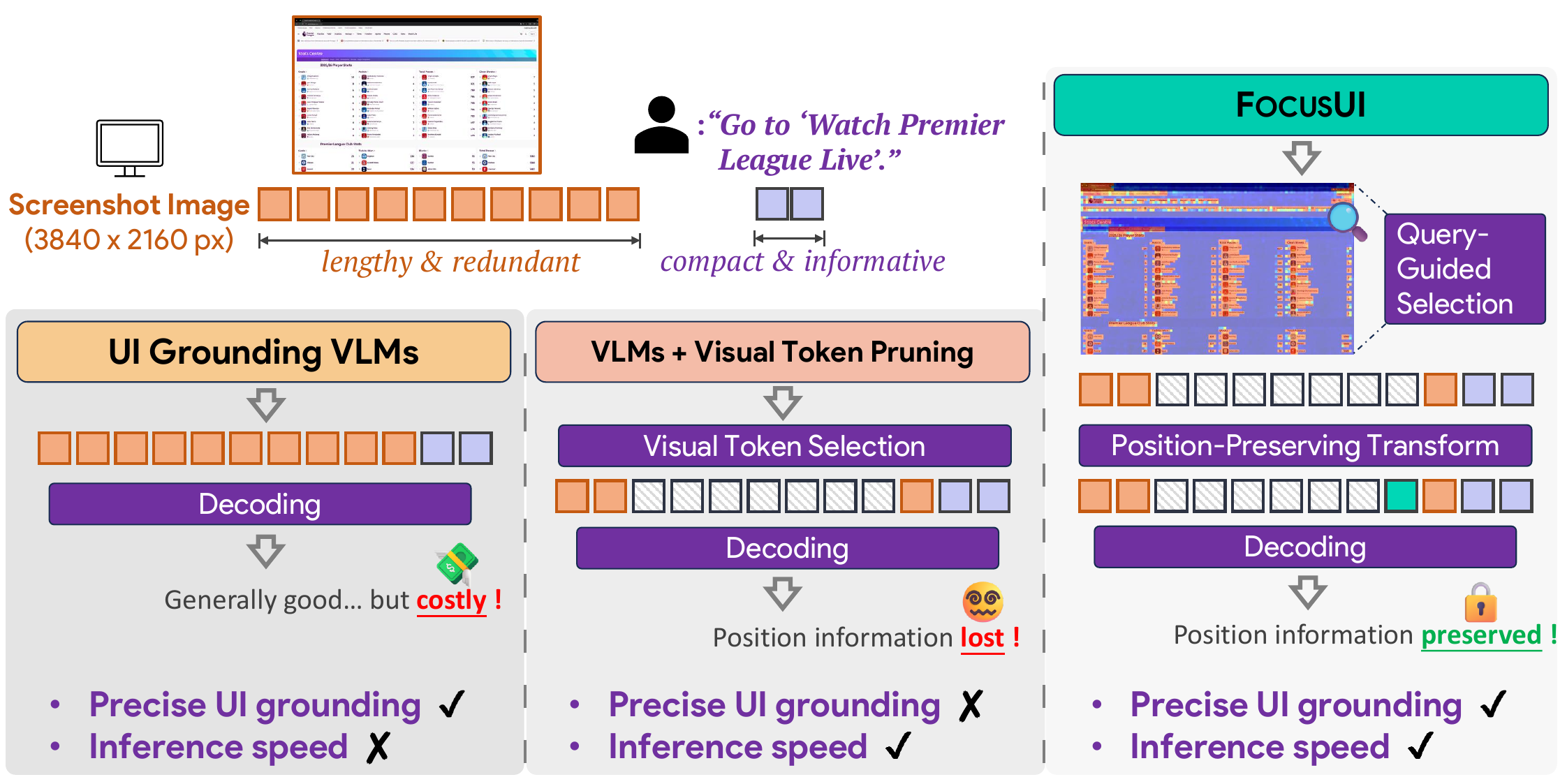}
    \vspace{-1.25em}
    \subcaption{\textbf{Comparison of vanilla UI grounding VLMs}, VLMs with visual token pruning, and our \focusui.}
    \vspace{0.1em}
    \label{fig:1a_acc_reduction}
    \end{subfigure}
    \begin{subfigure}[t]{0.492\linewidth}
    \centering
    \includegraphics[width=\linewidth]{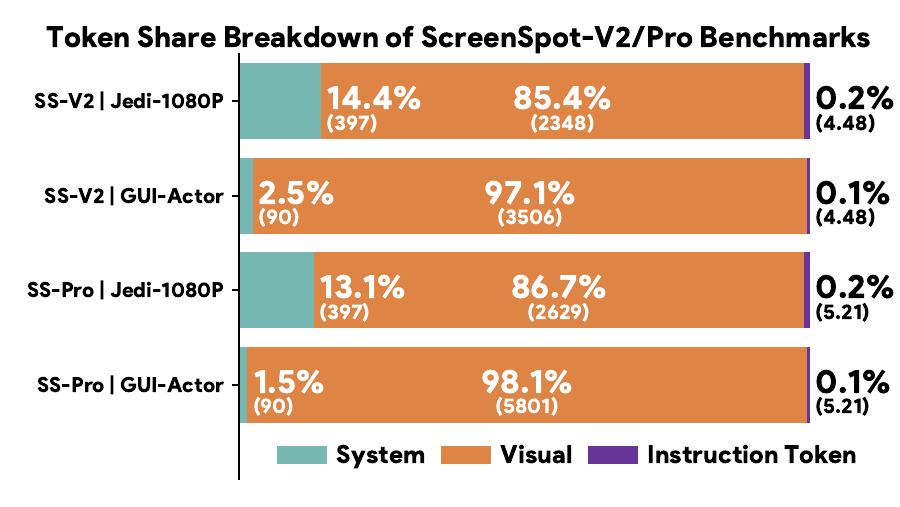}
    \captionsetup{margin={0pt,0.25em}}
    \subcaption{\textbf{Study 1:} The exceptionally high proportion of {\setlength{\fboxsep}{2.2pt}\colorbox[HTML]{E9AE87}{visual}} (screenshot) vs. {\setlength{\fboxsep}{2.2pt}\colorbox[HTML]{C5C8F0}{text}} (instruction) tokens in UI grounding tasks.}
    \label{fig:1b_token_share}
    \end{subfigure}
    \hfill
    \begin{subfigure}[t]{0.492\linewidth}
    \centering
    \includegraphics[width=\linewidth]{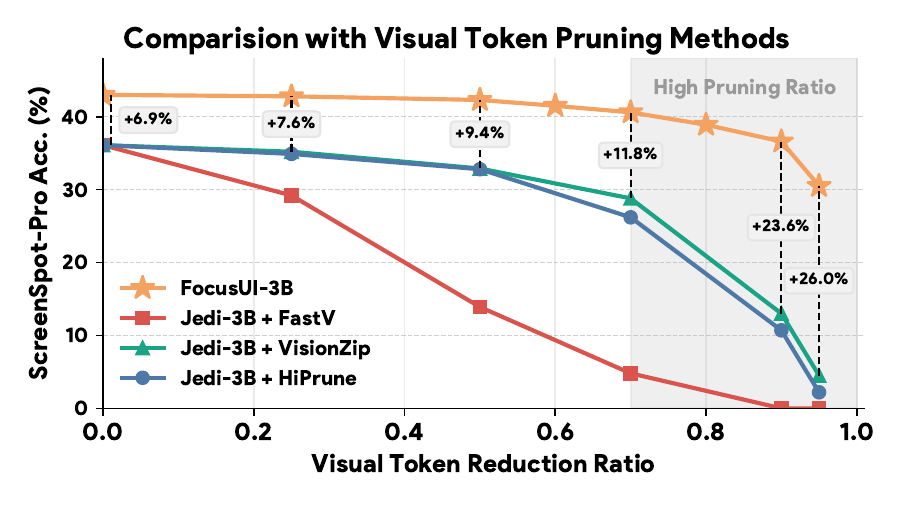}
    \captionsetup{margin={0.25em,0.0em}}
    \subcaption{\textbf{Study 2:} Our proposed position-preserving visual token selection vs. general visual token pruning methods.}
    \label{fig:1c_vs_pruning}
    \end{subfigure}
    \vspace{-0.5em}
    \caption{\focusui\ is an efficient UI grounding framework that selects \emph{instruction-relevant} visual tokens while \emph{preserving positional continuity}. \textbf{Study 1} provides motivation to address visual redundancy in UI grounding tasks, and \textbf{Study 2} demonstrates the effectiveness of the our position-preserving selection.}
    \vspace{-1.5em}
    \label{fig:1_teaser}
\end{figure}

\vspace{-0.5em}
\section{Introduction}
\label{sec:intro}
\vspace{-0.25em}

\begin{figure*}[!t]
    \centering
    \includegraphics[width=\linewidth]{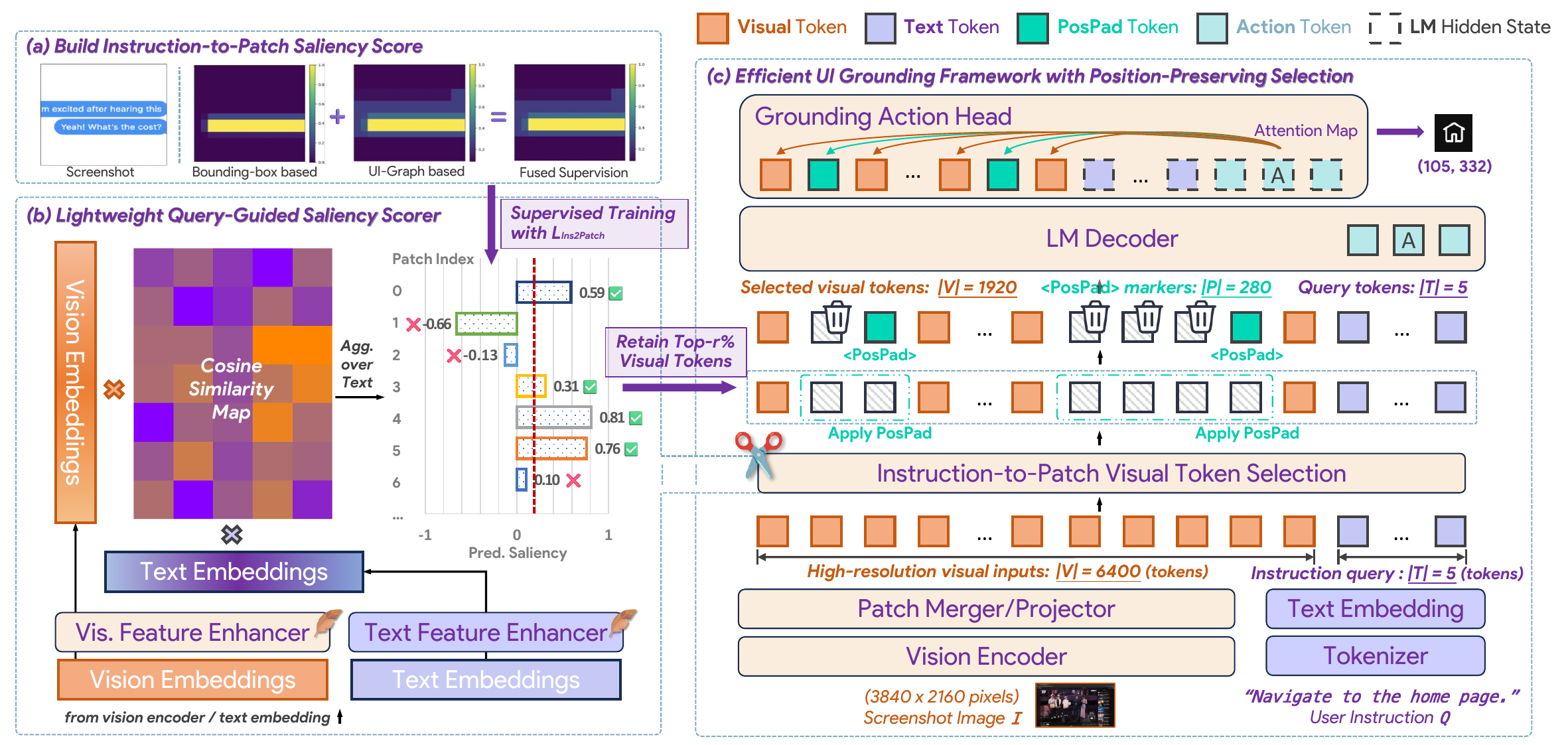}
    \vspace{-1.75em}
    \caption{\textbf{Overview of our proposed \focusui.} (a) Illustration of how the Instruction-to-Patch saliency score is constructed. (b) Query-guided Saliency Scorer and token selection. (c) Overall UI grounding framework illustrating how \pospad\ is applied to dropped sequences to preserve positional continuity. For clarity, we omit the system prompt in the token sequence.}
    \vspace{-0.75em}
    \label{fig:model_overview}
\end{figure*}

User interface (UI) visual grounding asks a model to locate a target region in a high-resolution screenshot given a natural language instruction.
Modern vision-language models (VLMs) have shown strong performance in UI tasks, including navigation and grounding, mainly driven by their abilities in processing high-resolution visual information. However, UI screenshots are typically high-resolution, and patchified into thousands of visual tokens that dominate the sequence budget (Fig.~\ref{fig:1b_token_share}).
This extreme visual token skew causes substantial computational overhead.
Although accuracy has improved rapidly, efficiency has been underexplored: na\"{\i}ve visual token pruning designed for natural images breaks \emph{positional continuity} in multimodal sequences and yields severe accuracy drops on precise UI grounding tasks. Recent studies in token pruning strategies aim to mitigate the rapidly growing computational cost by visual tokens. It is typically achieved by exploiting redundancy and importance variance, and applying selection in prefilling stage to reduce memory and computation costs during decoding. However, directly dropping visual tokens incurs position information loss, as sequence continuity is broken, leading to severe accuracy drops on precise UI grounding.

We present \focusui, an efficient UI grounding framework that selects \emph{instruction-relevant} visual tokens while preserving positional continuity needed for precise localization.
First, a \emph{lightweight} Query-Guided Saliency Scorer predicts per-patch relevance under dense supervision that fuses an instruction-conditioned bounding-box overlap signal with a rule-based UI-graph prior that down-weights large homogeneous regions.
Second, we apply \pospad\, which compacts each dropped \emph{contiguous} sequence into one learnable marker placed at the sequence’s last index, preserving its positional information.
This design mitigates sequence fragmentation and stabilizes grounding at aggressive retention ratios. 
\focusui\ integrates seamlessly with VLMs based on Qwen2.5-VL~\cite{bai2025qwen2_5vl} and Qwen3-VL~\cite{Qwen3-VL} of multiple sizes. Across experiments on four benchmarks, \focusui\ substantially speeds up inference and lowers peak GPU memory, while maintaining high accuracy.
The main contributions of this work include:
\begin{itemize}
    \item \textbf{Pioneering the task of efficient UI grounding.}
    We study the task characteristics and challenges of efficient UI grounding, presenting a dedicated approach that preserves accuracy while reducing visual tokens.
    \item \textbf{Instruction-to-patch selection with dense supervision.}
    We fuse a rule-based \emph{UI-graph} prior with instruction-conditioned bounding-box overlap to train a lightweight Query-Guided Saliency Scorer that predicts per-patch saliency and filters irrelevant tokens.
    \item \textbf{Position-preserving transformation.}
    We introduce \pospad\ to preserve sequence continuity during token selection, addressing the failure of general pruning methods on precise UI grounding tasks.
    \item \textbf{Practical integration and results.}
    We implement \focusui\ with Qwen2.5-VL and Qwen3-VL backbones of multiple sizes (2B, 3B, and 7B). Our models outperform the best previous state-of-the-art models and show good accuracy-efficiency trade-offs across four UI grounding benchmarks.
\end{itemize}


\section{Efficient UI Grounding: Task Characteristics and Challenges}
\label{sec:efficient-ui-grounding}

We identify two key challenges in UI grounding: \textbf{(1)} \textit{extreme token skew and redundancy from high-resolution screenshots}, and \textbf{(2)} \textit{accuracy collapse under na\"{\i}ve visual token pruning due to broken positional continuity}. In this section, we provide a comprehensive empirical analysis of these challenges, thereby elaborating on the motivation for our efficient UI grounding framework.

\subsection{High-Resolution Visual Understanding}
The task of UI grounding differs from natural visual understanding mainly in input characteristics: \textbf{UI screenshots are typically high resolution} (e.g., 2K at $2560\times1440$ or 4K at $3840\times2160$), \textbf{compositionally structured}, and \textbf{dominated by large homogeneous panes interspersed with small widgets}. To quantify this skewness, \textbf{Study 1} in Fig.~\ref{fig:1b_token_share} shows that visual (screenshot) tokens account for $\ge\!85.4\%$ of the tokens across two benchmarks and two grounding models, confirming a severe imbalance in visual tokens that incurs significant computational overhead.

This motivates an instruction-aware selection that prioritizes patches relevant to the instruction and de-emphasizes visually repetitive regions.
We implement this with an Instruction-to-Patch saliency score (\S\ref{sec:patch-scorer}) that fuses: (i) bounding-box overlap with ground-truth box and (ii) a rule-based UI-graph prior that down-weights large connected components, to guide the selection.

\subsection{Position Sensitivity in UI Grounding}
VLMs process multimodal inputs as an interleaved sequence of visual patch tokens and text tokens~\cite{su2024roformer}. In particular, Multimodal Rotary Position Embedding (M-RoPE)~\cite{wang2024qwen2vl} is designed for modeling spatial and temporal relationships. In practice, Qwen2-VL's M-RoPE decomposes rotary dimensions into temporal, height, and width components to encode a $(t,h,w)$ structure~\cite{huang2025revisitingmultimodalpositionalencoding}.
However, we find that \textbf{precise UI grounding is sensitive to the positional information of visual embeddings}, which makes token reduction more challenging. 
Direct pruning creates \emph{positional jumps} in the $(h,w)$ dimensions of M-RoPE sequence, leading to pronounced localization offsets on fine-grained targets.
To investigate this sensitivity, in \textbf{Study 2} of Fig.~\ref{fig:1c_vs_pruning}, we evaluate UI grounding models applied with advanced visual token pruning methods.
The sharp accuracy drop suggests that although these pruning methods work well for general visual understanding scenerios,  performance degrades dramatically on precise localization.

We address this with a \pospad\ (\S\ref{sec:last-pad}) strategy: for each \emph{contiguous} sequence of dropped visual tokens, we replace the sequence with a single learnable marker placed at the sequence’s \emph{last} index, inheriting that index’s $(h,w)$ positional information. This special marker preserves positional continuity and mitigates the disruption to the model's spatial understanding. Together, \textbf{Study 1} motivates \emph{what} to remove (instruction-irrelevant or homogeneous regions), and \textbf{Study 2} dictates \emph{how} to select (position-preserving rather than na\"{\i}ve dropping). These findings collectively form the motivation of our efficient UI grounding framework.


\section{\focusui}
\label{sec:method}
We introduce \focusui, a query-guided efficient UI grounding framework that selects instruction-relevant visual tokens while preserving positional continuity. 
As illustrated in Fig.~\ref{fig:model_overview}, \focusui\ comprises the following key components designed for efficient UI grounding: \textbf{(i)} a fused supervision of per-patch saliency score to identify instruction-relevant visual tokens, \textbf{(ii)} a lightweight Query-Guided Saliency Scorer for visual token selection, and \textbf{(iii)} a novel position-preserving \pospad\ strategy to preserve positional information during token selection. 
In the following sections, we introduce each component in detail.


\subsection{Instruction-to-Patch Saliency Score}

Motivated by observations in \S\ref{sec:efficient-ui-grounding}, we first construct dense supervision of per-patch saliency scores to select relevant visual tokens. We fuse two complementary components: (i) instruction-conditioned bounding-box overlap and (ii) a UI-graph prior via union-find that down-weights large homogeneous regions. 

\vspace{-1em}

{
\setlength{\algomargin}{0.2em} 
\begin{algorithm}[!b]
    \small
    \caption{Building Bounding-Box Saliency Score}\label{algo:bbox-scoring}
    \KwIn{$I\in[0,1]^{H\times W\times 3}$, patch size $p$, ground-truth bbox $b_{gt}=(x_1,y_1,x_2,y_2)$}
    \KwOut{$S_{\mathrm{bbox}}\in[0,1]^{G_h\times G_w}$}
    $G_h \gets \lfloor H/p \rfloor,\quad G_w \gets \lfloor W/p \rfloor$ \\
    \For{$i \gets 0$ \KwTo $G_h-1$}{
    \For{$j \gets 0$ \KwTo $G_w-1$}{
        $R_{i,j} \gets [jp,ip,(j{+}1)p,(i{+}1)p]$;\quad
        $S_{\mathrm{bbox}}[i,j] \gets \mathrm{area}(R_{i,j}\cap b_{gt})/p^2$
    }
    }
    \Return $S_{\mathrm{bbox}}$
\end{algorithm}
}
{
\setlength{\algomargin}{0.2em} 
\begin{algorithm}[!b]
    \small
    \caption{Building UI-Graph Saliency Score}\label{algo:uig-scoring}
    \KwIn{$I\in[0,1]^{H\times W\times 3}$, threshold $\tau$, patch size $p$} 
    \KwOut{$S_{\mathrm{uig}}\in[0,1]^{G_h\times G_w}$}
        $G_h \gets \lfloor H/p \rfloor,\quad G_w \gets \lfloor W/p \rfloor$ \\
        Form patch pixels ${PP}_{i,j}\in\mathbb{R}^{3\times p\times p}$ for $0\le i<G_h,\ 0\le j<G_w$ \\
        \textbf{Union-Find} on nodes $(i,j)$ 
        \For{$i \gets 0$ \KwTo $G_h-1$}{
        \For{$j \gets 0$ \KwTo $G_w-1$}{
            \If{$j+1 < G_w$ \textbf{and} $\lVert \mathrm{vec}({PP}_{i,j}) - \mathrm{vec}({PP}_{i,j+1}) \rVert_2 < \tau$}{
            \textsc{union}$\big((i,j),(i,j+1)\big)$
            }
            \If{$i+1 < G_h$ \textbf{and} $\lVert \mathrm{vec}({PP}_{i,j}) - \mathrm{vec}({PP}_{i+1,j}) \rVert_2 < \tau$}{
            \textsc{union}$\big((i,j),(i+1,j)\big)$
            }
        }
        }
        Obtain component ids $r_{i,j}\gets \textsc{find}(i,j)$ \\
        Counts $n_u \gets \big|\{(i,j): r_{i,j}=u\}\big|$ for each unique root $u$ \\
        \textbf{Assigning Weights:}\quad $w_u \gets \big(\max\{1,\ \ln(n_u+1)\}\big)^{-1}$ \\
        Set $S_{\mathrm{uig}}[i,j] \gets w_{r_{i,j}}$ for all $i,j$ \\
    \Return $S_{\mathrm{uig}}$
\end{algorithm}
}

\paragraph{Bounding-Box Saliency Score.}
As summarized in Alg.~\ref{algo:bbox-scoring},
we partition the image into a $G_h\times G_w$ patch grid with patch size $p$, and denote the patch cell by $R_{i,j}=[jp,ip,(j{+}1)p,(i{+}1)p]$.
Given an element bounding box $b_{gt}$, each patch cell receives a score proportional to its overlap with $b_{gt}$.
We set $S_{\mathrm{bbox}}\in[0,1]$ with normalized overlap $\mathrm{area}(R_{i,j}\cap b_{gt})/p^2$ so that fully covered patches score $1$ and disjoint patches score $0$, inducing a center-to-edge decay along the box boundary.

\paragraph{UI-Graph Saliency Score.}
To further suppress background regions and enrich supervision on non-annotated regions, we propose a UI-graph saliency score based on union–find over connected components of visual patches, which is inspired by the \emph{UI-graph} prior in ShowUI~\cite{lin2024showui}.
Specifically, we treat each patch $(i,j)$ in $R_{i,j}$ as a node and connect 4-neighborhood pairs whose $\ell_2$ distance in the RGB space is below a threshold $\tau$. Such union–find groups connected components whose size $n_u$ reflects how visually repetitive a region is.

We then assign a weight $w_u\!=\!(\max\{1,\ln(n_u\!+\!1)\})^{-1}$ to each patch so that large homogeneous regions (\eg, empty backgrounds) receive lower weights.
The UI-graph score $S_{\mathrm{uig}}$ sets each patch to its component weight $w_u$. Such design naturally suppresses background regions and enhances the saliency of distinctive elements.
This score is instruction-agnostic, annotation-free, and complements $S_{\mathrm{bbox}}$ for each patch.
See Alg.~\ref{algo:uig-scoring} for the full procedure.

\begin{figure}[!b]
    \vspace{-0.25em}
    \centering
    \includegraphics[width=\linewidth]{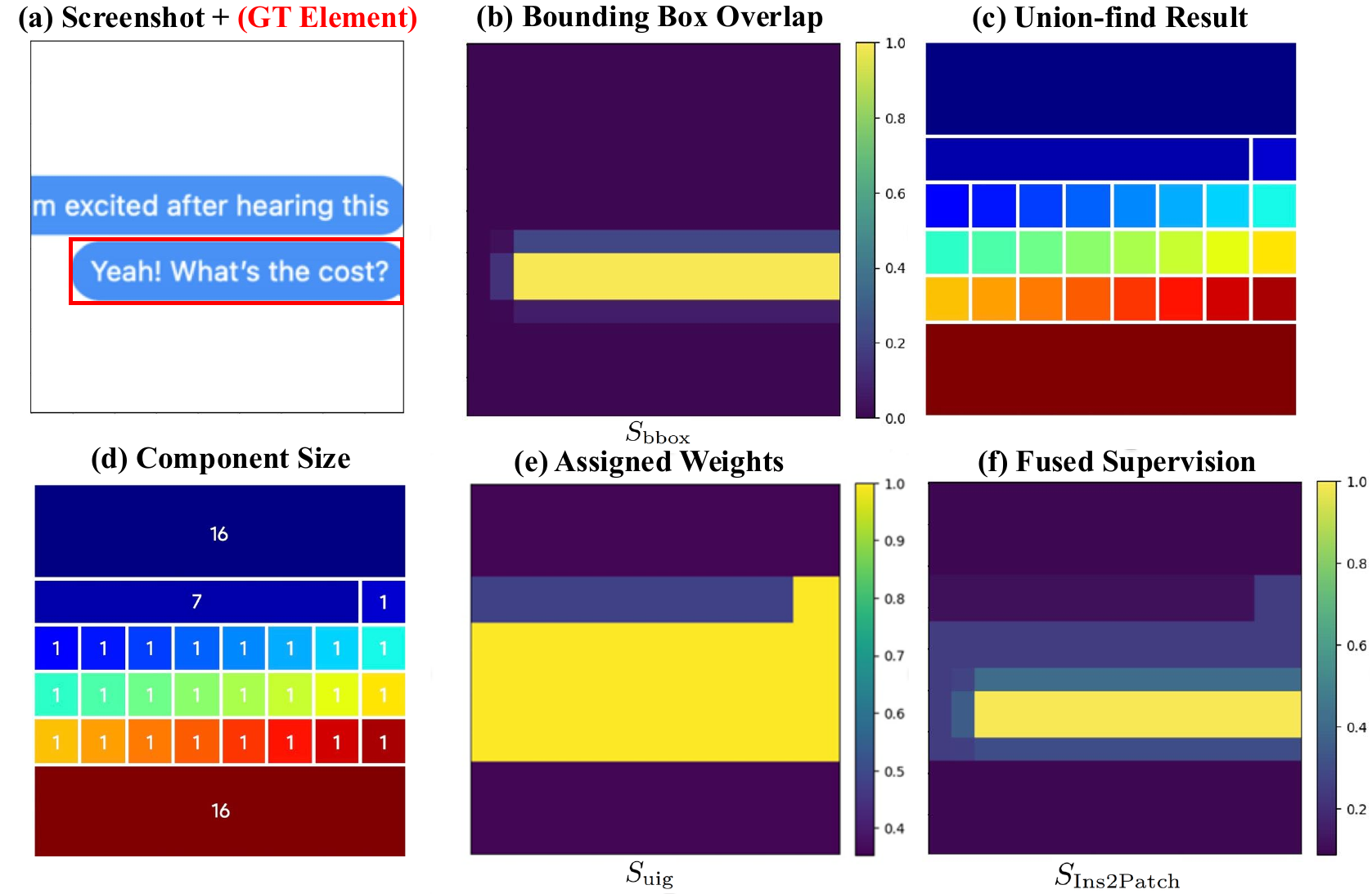}
    \caption{Illustrative example of building the Instruction-to-Patch saliency score. \textbf{(a)} Screenshot $I$ with ground-truth bounding box $b_{gt}$. \textbf{(b)} Bounding-box saliency score $S_{\mathrm{bbox}}$. \textbf{(c)} Union-find results. \textbf{(d)} Size of each connected component $n_u$. \textbf{(e)} UI-graph saliency score $S_{\mathrm{uig}}$. \textbf{(f)} Fused supervision $S_{\mathrm{Ins}2\mathrm{Patch}}$ by combining \textbf{(d)} and \textbf{(e)}. Brighter regions represent positive patches and darker regions represent negative patches.
    \vspace{-0.5em}
    }
    \label{fig:patchscore}
\end{figure}

\paragraph{Fuse Supervision.} 
Finally, we fuse the two scores to obtain joint supervision $S_{\mathrm{Ins}2\mathrm{Patch}}$ as Instruction-to-Patch saliency score:
\vspace{-0.5em}
\begin{equation}
    S_{\mathrm{Ins}2\mathrm{Patch}}=\lambda\,S_{\mathrm{bbox}}+(1-\lambda)\,S_{\mathrm{uig}}
    \label{eq:patchscore-fused}
    \vspace{-0.25em}
\end{equation}
where $\lambda\in[0,1]$ is a controllable weight and empirically set to $0.8$ across experiments. Fig.~\ref{fig:patchscore} provides an illustration of the two components and the final fused supervision.

\subsection{Lightweight Query-Guided Saliency Scorer}
\label{sec:patch-scorer}
With the obtained per-patch supervision $S_{\mathrm{Ins}2\mathrm{Patch}}$ from Eq.~\eqref{eq:patchscore-fused}, we train a \emph{lightweight} module, Query-Guided Saliency Scorer, that predicts per-patch saliency from similarities between patch and query text embeddings in the VLM backbone, as shown in Fig.~\ref{fig:model_overview} (b).

Concretely, let $\{v_i\}_{i=1}^{M}$ be patch embeddings from the vision encoder and $\{e_j\}_{j=1}^{N}$ be query text embeddings (only the part corresponding to the instruction) in the language model (LM) space. We use a self-attention layer to enhance features in each modality, preserving the original embedding semantics while strengthening cross-modal interactions. A tanh constraint followed by $\ell_2$ normalization is applied to each feature to bound the similarities.
We then compute token-wise similarities $\mathcal{P} \in \mathbb{R}^{M\times N}$ by a matrix product between patch and text embeddings. Finally, we aggregate the similarities over text query dimensions with mean pooling to get per-patch saliency scores $s_i$:
\vspace{-0.5em}
\begin{equation}
    \mathcal{P}  \,=\, \tilde V\tilde E^{\top} \in \mathbb{R}^{M\times N},\quad
    s_i \,=\, \frac{1}{N}\sum_{j=1}^{N} \mathcal{P}_{i,j} \,.
\label{eq:ins2patch-si}
\vspace{-0.5em}
\end{equation}

To train the Query-Guided Saliency Scorer, we convert scores to probabilities and optimize a KL divergence objective. Given fused supervision from Eq.~\eqref{eq:patchscore-fused}, we minimize:
\begin{equation}
    \mathcal{L}_{\text{Ins2Patch}}=\mathrm{KL}\!\left(\mathrm{softmax}\!\left(S_{\mathrm{Ins}2\mathrm{Patch}}\right)\,\middle\|\,\mathrm{softmax}\!\left(s\right)\right).
    \label{eq:loss-ps-impl}
\end{equation}

\subsection{\pospad: Positional Continuity Preservation}
\label{sec:last-pad}

\paragraph{Token Selection Policy.}

We first apply top-$K$ selection over predicted per-patch saliency scores $\{s_i\}_{i\in\mathcal{I}}$ from Eq.~\eqref{eq:ins2patch-si}. Given a retention ratio $r\in(0,1]$, the number of kept patches is set to $K=\lfloor rM\rfloor$. Let $\gamma$ be the $K$-th element of the sorted list $\{s_i\}_{i\in\mathcal{I}}$. We form the kept index set $\mathcal{K}=\{ i\in\mathcal{I}\mid s_i\ge\gamma \}$ and drop the remaining indices $\mathcal{D}=\{ i\in\mathcal{I}\mid s_i<\gamma \}$. 

\begin{figure}[!t]
    \centering
    \includegraphics[width=1.\linewidth]{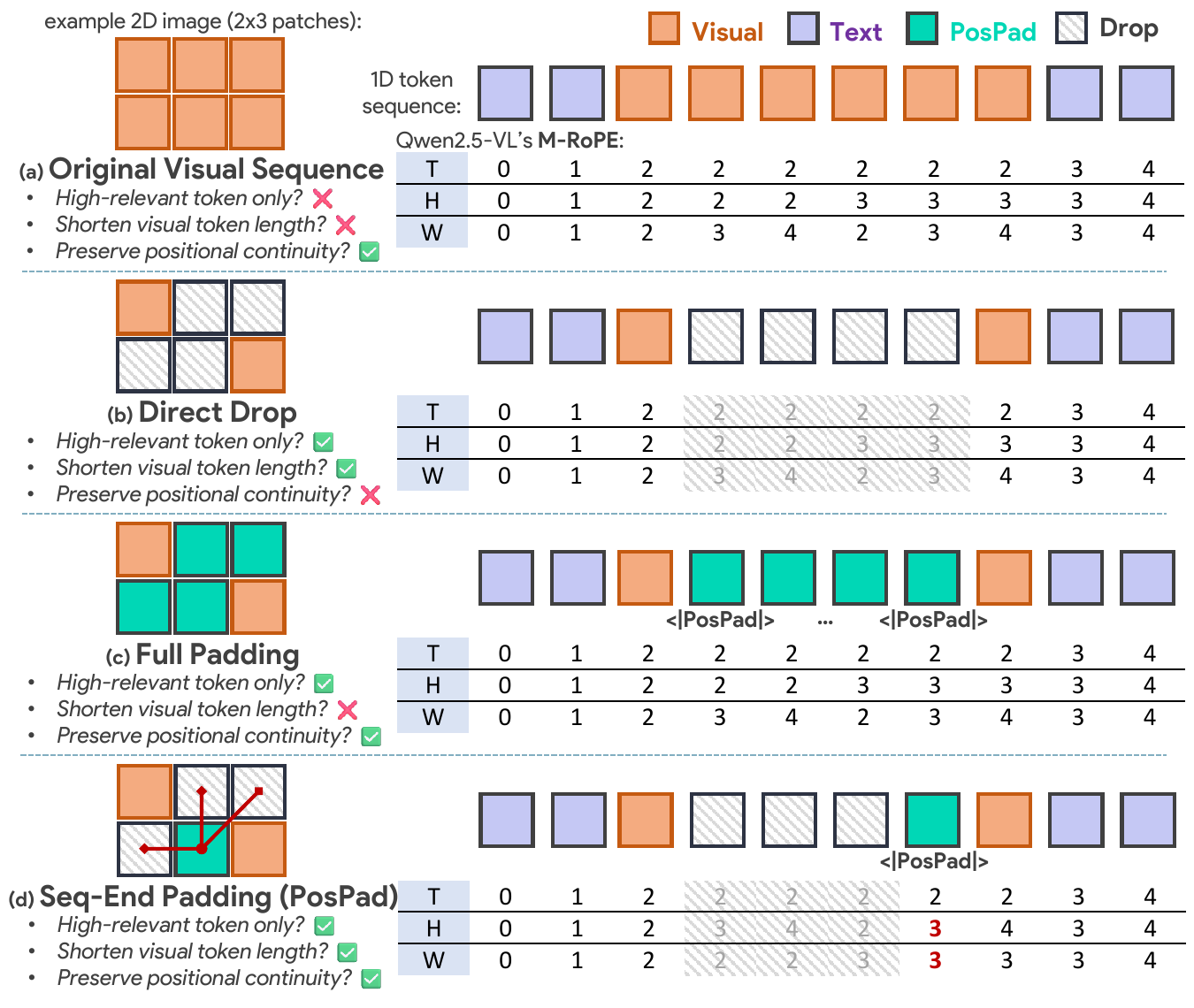}
    \vspace{-1.75em}
    \caption{Illustration of \pospad\ sequence transformation for positional continuity preservation via an example 2D image (2$\times$3 patches) and its 1D sequence. A learnable \pospadtoken\ marker is placed at the last index of each contiguous sequence of dropped visual tokens, as illustrated by strategy (d).}
    \vspace{-1.25em}
    \label{fig:pospad}
\end{figure}

\paragraph{Sequence Transformation.}
After selecting instruction-relevant visual tokens, we further refine the sequence to alleviate positional information loss in the model’s spatial understanding.
We introduce \pospad, a position-preserving sequence transformation that replaces each \emph{contiguous sequence} of dropped visual tokens with a \emph{single} learnable special token \pospad\ placed at the \emph{last index} of that sequence. The illustration of \pospad\ is shown in Fig.~\ref{fig:pospad}.

Specifically, given the original visual token sequence $x_{1:M}$, the kept index set $\mathcal{K}$, and the drop index set $\mathcal{D}$ defined above, we partition $\mathcal{D}$ into \emph{contiguous sequences} (\ie, \emph{maximal consecutive sequences}) $\mathcal{R}_1,\ldots,\mathcal{R}_U$ with respect to the 1D flattened sequence order.
For each sequence $\mathcal{R}_u$, we keep only its last index $r_u^{\mathrm{end}}=\max\,\mathcal{R}_u$ and remove the others.
Let $\mathcal{E}_{\text{seq-end}}=\{ r_u^{\mathrm{end}} \}_{u=1}^{U}$ denote the set of sequence-end indices, and define the preserved index set $\mathcal{S}=\mathcal{K}\,\cup\,\mathcal{E}_{\text{seq-end}}$.
We then replace each contiguous sequence with a single marker \pospadtoken\ and keep all other tokens unchanged:
\begin{equation}
    \vspace{-0.5em}
    \begin{aligned}
    &x'_j \,=\, \begin{cases}
    \pospadtoken & \text{if } j\in\mathcal{E}_{\text{seq-end}}, \\
    x_j & \text{if } j\in\mathcal{K},
    \end{cases} \\
    &\mathrm{\pospad}(x_{1:M}) \,=\, \{ x'_j \}_{j\in\mathcal{S}}\ .
    \end{aligned}
    \label{eq:last-pad}
\end{equation}
Thus, the final output length of visual tokens is $M' = M - (|\mathcal{D}| - U)$, with the total number of \pospadtoken\ tokens being $U$.
Each dropped sequence $\mathcal{R}_u$ reduces the sequence by $|\mathcal{R}_u| - 1$ while preserving positional continuity at the sequence end.
Concrete examples of $M$, $M'$, and $U$ under different retention ratios are investigated in Tab.~\ref{tab:rq5-ablation-ratios}.

Compared to direct dropping, \pospad\ preserves positional continuity and empirically stabilizes the model’s spatial understanding. Alternative strategies are also studied in \S\ref{sec:rq5-ablation}.
Since \pospad\ alters only sequence sparsity and not token indices or rotary bases, it is compatible with common M-RoPE implementations and requires no modifications to the downstream LM architecture.

\subsection{Efficient UI Grounding Framework}

\paragraph{Integration with VLMs.}
We integrate our visual token selection strategy into existing VLMs before visual patch embeddings are fed into the LM decoder. Concretely, the Query-Guided Saliency Scorer takes the patch features $\{v_i\}_{i=1}^{M}$ and the instruction token embeddings ${e_j}$, computes scores ${s_i}$ via Eq.~\eqref{eq:ins2patch-si}, and selects the top-$K$ indices $\mathcal{K}$ for a given retention ratio $r$. We then refine the sequence with \pospad, yielding a compact visual sequence of length $M'\ll M$ that preserves positional continuity. The LM decoder processes this sequence without altering its original architecture. We apply our framework to Qwen2.5-VL and Qwen3-VL models. For the Qwen3-VL model with a DeepStack~\cite{meng2024deepstack} vision encoder, deep visual embeddings are gathered only for the kept image tokens $\mathcal{K}$.

\paragraph{Coordinate-free UI Grounding with Selected Patches.}
We find the coordinate-free UI grounding scheme from GUI-Actor~\cite{wu2025gui} most compatible with our selection: the model grounds elements directly at the patch embeddings with an extra action head on top of the LM decoder, while our visual token selection reduces candidates by discarding instruction-irrelevant regions. Specifically, the decoder $\mathrm{LM}$ outputs a sequence of action tokens:
\begin{equation}
    \begin{aligned}
    \vspace{-0.5em}
    \mathrm{LM}(I,q) = \{&x_{1:i-1}, \texttt{<ACTOR\_START>}, \\
                        &\texttt{<ACTOR>}, \texttt{<ACTOR\_END>}, x_{i+3:N}\}.
    \end{aligned}
    \vspace{-0.5em}
\label{eq:lm-out}
\end{equation}
Then the action head aligns $h_{\texttt{ACTOR}}$ with visual patches to produce an attention map over patches. We first contextually refine selected patch features $\{\tilde v_i\}_{i=1}^{M'}$ with a self-attention layer:
$\tilde v_1,\ldots,\tilde v_{M'}=\mathrm{SelfAttn}(v_1,\ldots,v_{M'}).$
Then we project $h_{\texttt{ACTOR}}$ and each $\tilde v_i$ with separate $\mathrm{MLP}_{T}$ and $\mathrm{MLP}_{V}$ and compute attention scores:
\begin{equation}
    \begin{aligned}
    \vspace{-0.5em}
    &z=\mathrm{MLP}_{T}(h_{\texttt{ACTOR}}), \quad z_i=\mathrm{MLP}_{V}(\tilde v_i), \\
    &\alpha_i=\tfrac{z^\top z_i}{\sqrt{d}}, \quad a_i=\mathrm{softmax}(\alpha)_i.
    \end{aligned}
    \vspace{-0.5em}
\label{eq:attn}
\end{equation}
The distribution ${a_i}$ identifies the most relevant regions for executing the action. With selected visual tokens, such an action head benefits from fewer visual candidates and retained patches that are more relevant to the instruction. 

\paragraph{Training Objective.}
The Query-Guided Saliency Scorer is trained end-to-end with the downstream LM objective next-token prediction loss $\mathcal{L}_\text{NTP}$ and an action-attention loss $\mathcal{L}_\text{Attn}$ for grounding:
\vspace{-0.25em}
\begin{equation}
    \mathcal{L}_{\text{Attn}}\!=\!\sum_{i=1}^{M'} p_i \log\!\frac{p_i}{a_i},\;
    p_i\!=\!\frac{y_i}{\sum_{j=1}^{M'} y_j\!+\!\epsilon},\;
    i\!=\!1,\!\dots,\!M'
\end{equation}
where $y_i$ denotes the attention score label for the $i$-th patch ($1$ if it overlaps with the ground-truth bounding box, $0$ otherwise) and $\epsilon$ is a small constant for numerical stability. The overall training objective is:
\begin{equation}
\vspace{-0.5em}
\mathcal{L}=\mathcal{L}_{\text{Ins2Patch}}+\mathcal{L}_\text{NTP}+\mathcal{L}_\text{Attn}.
\label{eq:loss-total}
\vspace{-0.25em}
\end{equation}

\begin{table*}[!t]
\centering
\small
\resizebox{\textwidth}{!}{
    \setlength{\tabcolsep}{1.85pt}
    \begin{tabular}{lccccccc@{\hspace{4pt}}||ccccccccccc}
    \toprule
    \multirow{2}{*}{\textbf{Model}} & \multicolumn{7}{c}{\textbf{ScreenSpot-V2}} & \multicolumn{9}{c}{\textbf{ScreenSpot-Pro}} \\
    \cmidrule(lr){2-8} \cmidrule(lr){9-17}
    & \textbf{Mob.-T} & \textbf{Mob.-I} & \textbf{Des.-T} & \textbf{Des.-I} & \textbf{Web-T} & \textbf{Web-I} & \cellcolor{niceorange}\textbf{Avg} & \textbf{Dev} & \textbf{Cre.} & \textbf{CAD} & \textbf{Sci.} & \textbf{Office} & \textbf{OS} & \textbf{Avg-T} & \textbf{Avg-I} & \cellcolor{niceorange}\textbf{Avg} \\
    \midrule
    Operator~\cite{openai2025operator} & 47.3 & 41.5 & 90.2 & 80.3 & 92.8 & 84.3 & \cellcolor{niceorange}70.5 & 35.1 & 39.6 & 16.1 & 43.7 & 53.0 & 32.7 & 45.0 & 23.0 & \cellcolor{niceorange}36.6 \\
    OS-Atlas-7B~\cite{wu2024atlas} & 95.2 & 75.8 & 90.7 & 63.6 & 90.6 & 77.3 & \cellcolor{niceorange}84.1 & 17.7 & 17.9 & 10.3 & 24.4 & 27.4 & 16.8 & 28.1 & 4.0 & \cellcolor{niceorange}18.9 \\
    Aguvis-7B~\cite{xu2024aguvis} & 95.5 & 77.3 & 95.4 & 77.9 & 91.0 & 72.4 & \cellcolor{niceorange}86.0 & 16.1 & 21.4 & 13.8 & 34.6 & 34.3 & 19.4 & - & - & \cellcolor{niceorange}22.9 \\
    Tong-UI-7B~\cite{zhang2025tongui} & 93.1 & 81.5 & 96.4 & 82.9 & 90.2 & 84.7 & \cellcolor{niceorange}88.7 & 22.7 & 21.1 & 15.3 & 34.3 & 38.3 & 18.4 & 35.1 & 8.0 & \cellcolor{niceorange}25.7 \\
    UGround-V1-7B~\cite{gou2024navigating} & 95.0 & 83.3 & 95.0 & 77.8 & 92.1 & 77.2 & \cellcolor{niceorange}87.6 & 28.1 & 31.7 & 14.6 & 39.0 & 49.6 & 24.5 & - & - & \cellcolor{niceorange}31.1 \\
    UI-TARS-7B~\cite{qin2025ui} & 96.9 & 89.1 & 95.4 & 85.0 & 93.6 & 85.2 & \cellcolor{niceorange}91.6 & 36.1 & 32.8 & 18.0 & 50.0 & 53.5 & 24.5 & 47.8 & 16.2 & \cellcolor{niceorange}35.7 \\
    UI-TARS-72B~\cite{qin2025ui} & 94.8 & 86.3 & 91.2 & 87.9 & 91.5 & 87.7 & \cellcolor{niceorange}90.3 & 40.8 & 39.6 & 17.2 & 45.7 & 54.8 & 30.1 & 50.9 & 17.5 & \cellcolor{niceorange}38.1 \\
    UI-TARS-1.5-7B~\cite{qin2025ui} & - & - & - & - & - & - & \cellcolor{niceorange}90.0 & 31.8 & 40.2 & 31.8 & 47.2 & 65.6 & 33.2 & - & - & \cellcolor{niceorange}42.6\\
    Qwen2.5-VL-3B~\cite{bai2025qwen2_5vl} & 93.4 & 73.5 & 88.1 & 58.6 & 88.0 & 71.4 & \cellcolor{niceorange}80.9 & 21.4 & 25.8 & 18.4 & 29.5 & 40.9 & 20.4 & 37.8 & 6.6 & \cellcolor{niceorange}25.9 \\
    Qwen2.5-VL-7B~\cite{bai2025qwen2_5vl} & 97.6 & 87.2 & 90.2 & 74.2 & 93.2 & 81.3 & \cellcolor{niceorange}88.8 & 29.1 & 24.9 & 13.8 & 31.1 & 45.7 & 22.4 & 39.9 & 7.6 & \cellcolor{niceorange}27.6 \\
    Qwen2.5-VL-32B~\cite{bai2025qwen2_5vl} & 97.9 & 88.2 & 98.5 & 79.3 & 91.2 & 86.2 & \cellcolor{niceorange}91.3 & 48.5 & 41.1 & 32.6 & 57.1 & 67.4 & 42.3 & 63.2 & 22.5 & \cellcolor{niceorange}47.6 \\
    GUI-Actor-3B~\cite{wu2025gui} & 97.6 & 83.4 & 96.9 & 83.6 & 94.0 & 85.7 & \cellcolor{niceorange}91.0 & 39.8 & 36.7 & 34.1 & 49.6 & 61.3 & 35.2 & - & - & \cellcolor{niceorange}42.2 \\
    GUI-Actor-7B~\cite{wu2025gui} & 97.6 & 88.2 & 96.9 & 85.7 & 93.2 & 86.7 & \cellcolor{niceorange}92.1 & 38.1 & 41.4 & 38.3 & 50.8 & 63.0 & 38.8 & - & - & \cellcolor{niceorange}44.6 \\
    Jedi-3B~\cite{xie2025scaling} & 96.6 & 81.5 & 96.9 & 78.6 & 88.5 & 83.7 & \cellcolor{niceorange}88.6 & 38.1 & 34.6 & 23 & 38.6 & 57.0 & 25.0 & 49.8 & 13.7 & \cellcolor{niceorange}36.1 \\
    Jedi-7B~\cite{xie2025scaling} & 96.9 & 87.2 & 95.9 & 87.9 & 94.4 & 84.2 & \cellcolor{niceorange}91.7 & 27.4 & 34 & 32.2 & 52.4 & 68.7 & 26.0 & 52.6 & 18.2 & \cellcolor{niceorange}39.5 \\
    \midrule
    \rowcolor{ratioHundred}\focusuithreebillion\ {\footnotesize (\tighttt{$r=100\%$})} & 99.2 & 85.9 & 96.1 & 87.3 & 95.4 & 81.9 & 91.5 & 43.1 & 37.0 & 37.6 & 48.4 & 61.7 & 38.3 & 59.3 & 18.9 & 43.8 \\
    \rowcolor{ratioFifty}\focusuithreebillion\ {\footnotesize (\tighttt{$r=50\%$})} & 98.8 & 86.9 & 95.0 & 87.3 & 95.4 & 81.9 & 91.4 & 42.1 & 37.0 & 36.4 & 46.9 & 58.3 & 35.2 & 56.7 & 19.0 & 42.3 \\
    \rowcolor{ratioThirty}\focusuithreebillion\ {\footnotesize (\tighttt{$r=30\%$})} & 98.5 & 85.3 & 96.1 & 87.3 & 94.3 & 81.9 & 91.0 & 38.1 & 35.8 & 33.3 & 44.5 & 57.8 & 37.2 & 55.0 & 17.4 & 40.6 \\
    \midrule
    \rowcolor{ratioHundred}\focusuisevenbillion\ {\footnotesize (\tighttt{$r=100\%$})} & 98.8 & 91.6 & 95.6 & 92.1 & 95.0 & 84.4 & 93.1 & 44.5 & 41.1 & 42.9 & 52.0 & 69.6 & 44.4 & 64.7 & 21.9 & 48.3 \\
    \rowcolor{ratioFifty}\focusuisevenbillion\ {\footnotesize (\tighttt{$r=50\%$})} & 98.8 & 92.2 & 93.9 & 87.3 & 95.0 & 85.2 & 92.6 & 42.8 & 40.5 & 40.2 & 51.6 & 67.0 & 40.3 & 61.7 & 21.9 & 46.5 \\
    \rowcolor{ratioThirty}\focusuisevenbillion\ {\footnotesize (\tighttt{$r=30\%$})} & 98.8 & 90.1 & 93.3 & 85.7 & 93.9 & 85.2 & 91.8 & 38.8 & 39.9 & 42.9 & 49.2 & 64.4 & 38.8 & 60.4 & 20.4 & 45.1 \\
    \bottomrule
    \end{tabular}
}
\vspace{-5pt}
\caption{Performance comparison on \textit{ScreenSpot-V2}~\cite{wu2024atlas} and \textit{ScreenSpot-Pro}~\cite{li2025screenspot}.}
\vspace{-10pt}
\label{tab:screenspot_v2_pro}
\end{table*}

\vspace{-0.25em}
\section{Experiments}
\label{sec:experiments}

\subsection{Experimental Setup}

\begin{table}[!ht]
\centering
\small
\resizebox{\columnwidth}{!}{
\setlength{\tabcolsep}{1.8pt}
\begin{tabular}{lcccccc}
  \toprule
  \textbf{Model} & \textbf{Text} & \textbf{Elem} & \textbf{Layout} & \textbf{Manip} & \textbf{Refuse} & \cellcolor{niceorange}\textbf{Avg} \\
  \midrule
  Gemini-2.5-Pro~\cite{google2024gemini}     & 59.8 & 45.5 & 49.0 & 33.6 & 38.9 & \cellcolor{niceorange}45.2 \\
  Operator~\cite{openai2025operator}           & 51.3 & 42.4 & 46.6 & 31.5 & 0.0  & \cellcolor{niceorange}40.6 \\
  UGround-V1-7B~\cite{gou2024navigating}      & 51.3 & 40.3 & 43.5 & 24.8 & 0.0  & \cellcolor{niceorange}36.4 \\
  Aguvis-7B~\cite{xu2024aguvis}          & 55.9 & 41.2 & 43.9 & 28.2 & 0.0  & \cellcolor{niceorange}38.7 \\
  UI-TARS-7B~\cite{qin2025ui}         & 60.2 & 51.8 & 54.9 & 35.6 & 0.0  & \cellcolor{niceorange}47.5 \\
  UI-TARS-1.5-7B~\cite{qin2025ui}     & 70.1 & 57.9 & 59.7 & 51.7 & 0.0 & \cellcolor{niceorange}56.0 \\
  Qwen2.5-VL-3B~\cite{bai2025qwen2_5vl}      & 41.4 & 28.8 & 34.8 & 13.4 & 0.0  & \cellcolor{niceorange}27.3 \\
  Qwen2.5-VL-7B~\cite{bai2025qwen2_5vl}      & 45.6 & 32.7 & 41.9 & 18.1 & 0.0  & \cellcolor{niceorange}31.4 \\
  GUI-Actor-3B~\cite{wu2025gui}     & 60.5 & 56.1 & 58.5 & 32.2 & 0.0  & \cellcolor{niceorange}50.5 \\
  GUI-Actor-7B~\cite{wu2025gui}     & 60.2 & 54.2 & 58.1 & 30.9 & 0.0  & \cellcolor{niceorange}49.5 \\
  Jedi-3B~\cite{xie2025scaling}            & 67.4 & 53.0 & 53.8 & 44.3 & 7.4  & \cellcolor{niceorange}50.9 \\
  Jedi-7B~\cite{xie2025scaling}            & 65.9 & 55.5 & 57.7 & 46.9 & 7.4  & \cellcolor{niceorange}54.1 \\
  \midrule
  \rowcolor{ratioHundred}\focusuithreebillion\ {\footnotesize (\tighttt{$r=100\%$})} & 65.9 & 57.6 & 59.7 & 37.6 & 0.0 & 53.4 \\
  \rowcolor{ratioFifty}\focusuithreebillion\ {\footnotesize (\tighttt{$r=50\%$})} & 64.8 & 59.4 & 63.6 & 37.6 & 0.0 & 54.6 \\
  \rowcolor{ratioThirty}\focusuithreebillion\ {\footnotesize (\tighttt{$r=30\%$})} & 62.5 & 56.7 & 62.9 & 33.6 & 0.0 & 51.8\\
  \midrule
  \rowcolor{ratioHundred}\focusuisevenbillion\ {\footnotesize (\tighttt{$r=100\%$})} & 63.6 & 61.2 & 63.6 & 34.9 & 0.0  & 54.4 \\
  \rowcolor{ratioFifty}\focusuisevenbillion\ {\footnotesize (\tighttt{$r=50\%$})} & 64.0 & 62.1 & 63.6 & 31.5 & 0.0 & 54.1 \\
  \rowcolor{ratioThirty}\focusuisevenbillion\ {\footnotesize (\tighttt{$r=30\%$})} & 63.6 & 60.9 & 64.4 & 31.5 & 0.0 & 53.9 \\
  \bottomrule
\end{tabular}
}
\vspace{-0.5em}
\caption{Performance comparison on OSWorld-G~\cite{xie2025scaling}.}
\vspace{-1em}
\label{tab:osworld-g}
\vspace{-0.5em}
\end{table}

\begin{table}[!tbp]
\centering
\small
\resizebox{\columnwidth}{!}{  
\setlength{\tabcolsep}{5pt}
\begin{tabular}{lcccccc} 
\toprule
\textbf{Model} & \textbf{Basic} & \textbf{Functional} & \textbf{Spatial} & \cellcolor{niceorange}\textbf{Avg} \\
\midrule
Claude-3.7-Sonnet~\cite{anthropic2025claude} & 9.48 & 7.73 & 7.60 & \cellcolor{niceorange}8.27 \\
ShowUI-2B~\cite{lin2024showui} & 8.07 & 7.67 & 2.07 & \cellcolor{niceorange}5.94 \\
OSAtlas-7B~\cite{wu2024atlas} & 12.2 & 11.2 & 3.67 & \cellcolor{niceorange}9.02 \\
UGround-7B~\cite{gou2024navigating} & 11.5 & 12.2 & 2.79 & \cellcolor{niceorange}8.83 \\
UGround-V1-7B~\cite{gou2024navigating} & 15.4 & 17.1 & 6.25 & \cellcolor{niceorange}12.9 \\
Aguvis-7B~\cite{xu2024aguvis} & 17.8 & 18.3 & 5.06 & \cellcolor{niceorange}13.7 \\
UI-TARS-7B~\cite{qin2025ui} & 20.1 & 24.3 & 8.37 & \cellcolor{niceorange}17.6 \\
UI-TARS-72B~\cite{qin2025ui} & 31.4 & 30.5 & 14.7 & \cellcolor{niceorange}25.5 \\
GUI-Actor-3B~\cite{wu2025gui} & 27.4 & 24.6 & 7.0 & \cellcolor{niceorange}19.3 \\
GUI-Actor-7B~\cite{wu2025gui} & 30.1 & 28.1 & 7.8 & \cellcolor{niceorange}21.6 \\
Jedi-3B~\cite{xie2025scaling} & 22.3 & 25.2 & 9.35 & \cellcolor{niceorange}18.7 \\
Jedi-7B~\cite{xie2025scaling} & 32.3 & 30.5 & 12.8 & \cellcolor{niceorange}24.8 \\
\midrule
\rowcolor{ratioHundred}\focusuithreebillion\ {\footnotesize (\tighttt{$r=100\%$})} & 30.0 & 26.9&8.7&21.5 \\
\rowcolor{ratioFifty}\focusuithreebillion\ {\footnotesize (\tighttt{$r=50\%$})} & 29.7 & 26.0 & 8.2 & 20.9 \\
\rowcolor{ratioThirty}\focusuithreebillion\ {\footnotesize (\tighttt{$r=30\%$})} & 29.1 & 26.4 & 7.6 & 20.6 \\
\midrule
\rowcolor{ratioHundred}\focusuisevenbillion\ {\footnotesize (\tighttt{$r=100\%$})} & 33.6&31.2&11.2&24.9 \\
\rowcolor{ratioFifty}\focusuisevenbillion\ {\footnotesize (\tighttt{$r=50\%$})} & 32.5 & 31.0 & 11.3 & 24.5 \\
\rowcolor{ratioThirty}\focusuisevenbillion\ {\footnotesize (\tighttt{$r=30\%$})} & 32.3 & 29.2 & 11.0 & 23.8 \\
\bottomrule
\end{tabular}
}
\vspace{-0.5em}
\caption{Performance comparison on UI-Vision~\cite{nayak2025uivisiondesktopcentricguibenchmark}.}
\label{tab:uivision}
\vspace{-1.5em}
\end{table}

\paragraph{Implementation Details}

We adopt the state-of-the-art VLMs Qwen2.5-VL~\cite{bai2025qwen2_5vl} and Qwen3-VL~\cite{Qwen3-VL} as our base models, with different sizes to demonstrate the generalizability of our approach. We conduct supervised fine-tuning to obtain the following variants: \focusuithreebillion\ and \focusuisevenbillion\ with Qwen2.5-VL and \focusuitwobillionqwenthreevl\ with Qwen3-VL.

For fair comparison, we align the training budget with the baseline method GUI-Actor~\cite{wu2025gui}, using approximately 1M screenshots collected from several public UI datasets.
To ensure annotation quality, we follow V2P~\cite{chen2025v2pbackgroundsuppressioncenter} to apply OmniParser~\cite{lu2024omniparser} to filter samples whose IoU between ground-truth and detected boxes is below 0.3.
The visual token retention ratio $r$ is sampled uniformly from $(0.1, 1.0)$ during training.
All models are trained with DeepSpeed~\cite{10.1145/3394486.3406703} Zero-2 on $8\times$NVIDIA H200 GPUs for 1 epoch.
More training details are provided in the Appendix.

\paragraph{Evaluation Benchmarks}
We conduct experiments on four UI grounding benchmarks, including ScreenSpot-V2~\cite{wu2024atlas}, ScreenSpot-Pro~\cite{li2025screenspot}, OS-World-G~\cite{xie2025scaling}, and UI-Vision~\cite{nayak2025uivisiondesktopcentricguibenchmark}.
Among them, ScreenSpot-Pro features higher-resolution interfaces that simulate multi-source real-world applications, serving as a practical testbed for evaluating the properties of efficiency and precise UI grounding.

\begin{table}[!t]
\centering
\small
\resizebox{\columnwidth}{!}{  
\setlength{\tabcolsep}{2pt}
\begin{tabular}{lccc@{\hspace{4pt}}||ccc}
\toprule
\multirow{2}{*}{\textbf{Model}} & \multicolumn{3}{c}{\textbf{ScreenSpot-V2}} & \multicolumn{3}{c}{\textbf{ScreenSpot-Pro}} \\
\cmidrule(lr){2-4} \cmidrule(lr){5-7}  & \textbf{Avg-T} & \textbf{Avg-I} & \cellcolor{niceorange}\textbf{Avg} & \textbf{Avg-T} & \textbf{Avg-I} & \cellcolor{niceorange}\textbf{Avg} \\
\midrule
Qwen3-VL-2B$^{\dagger}$~\cite{bai2025qwen2_5vl} & 94.7 & 78.9 & \cellcolor{niceorange}87.8 & 52.8 & 16.7 & \cellcolor{niceorange} 39.0 \\
\midrule
\rowcolor{ratioHundred}\focusuitwobillionqwenthreevl\ {\footnotesize (\tighttt{$r=100\%$})} & 95.8 & 85.6 & 91.4 & 51.5 & 20.9 & 39.8 \\
\rowcolor{ratioFifty}\focusuitwobillionqwenthreevl\ {\footnotesize (\tighttt{$r=50\%$})} & 95.7 & 85.0 & 91.0 & 52.5 & 20.9 & 40.4 \\
\rowcolor{ratioThirty}\focusuitwobillionqwenthreevl\ {\footnotesize (\tighttt{$r=30\%$})} & 93.5 & 84.3 & 89.5 & 49.7 & 20.2 & 38.5 \\
\bottomrule
\end{tabular}
}
\vspace{-0.5em}
\caption{Performance comparison of models based on the \textbf{Qwen3-VL} backbone. $^{\dagger}$ indicates results obtained from our own evaluation of the official model on HuggingFace~\cite{wolf2020huggingfacestransformersstateoftheartnatural}.}
\label{tab:qwen3vl}
\vspace{-1.25em}
\end{table}

\subsection{Main Results}
\label{sec:results}

\noindent We organize our main results with the following five research questions (RQs):
\begin{itemize}
    \item \textbf{RQ1 (\S\ref{sec:rq1-performance} Performance)} : \textit{Can \focusui\ effectively reduce visual tokens while preserving accuracy?}
    \item \textbf{RQ2 (\S\ref{sec:rq2-pruning} Comparison to General Pruning Methods)}: \textit{How does \focusui\ compare to general visual token pruning methods?}
    \item \textbf{RQ3 (\S\ref{sec:rq3-efficiency} Efficiency Analysis)}: \textit{What efficiency gains does \focusui\ achieve under different settings?}
    \item \textbf{RQ4 (\S\ref{sec:rq4-qualitative} Qualitative Results)}: \textit{How does \focusui\ select instruction-relevant visual tokens?}
    \item \textbf{RQ5 (\S\ref{sec:rq5-ablation} Ablation Study)}: \textit{How do the components and retention settings affect performance?}
\end{itemize}

\subsubsection{RQ1: Performance}
\label{sec:rq1-performance}

Tables~\ref{tab:screenspot_v2_pro}, \ref{tab:osworld-g} and \ref{tab:uivision} report grounding performance on ScreenSpot-V2~\cite{wu2024atlas} \& ScreenSpot-Pro~\cite{li2025screenspot}, OS-World-G~\cite{xie2025scaling}, and UI-Vision~\cite{nayak2025uivisiondesktopcentricguibenchmark}, respectively.
We test a series of retention ratios $r\in\{100\%,50\%,30\%\}$ to characterize degradation curves and compare to dense baselines that consume all visual tokens. 
Across all four benchmarks, FOCUSUI exceeds GUI-specific baselines with the same size even at $30-50\%$ token retention, achieving state-of-the-art grounding performance.
Additionally, we report the performance of \focusuitwobillionqwenthreevl\ based on the more recent state-of-the-art Qwen3-VL~\cite{Qwen3-VL} backbone in Tab.~\ref{tab:qwen3vl}.
More detailed breakdown and retention ratio results are provided in the Appendix.

\subsubsection{RQ2: Comparison to General Pruning Methods}
\label{sec:rq2-pruning}

Tab.~\ref{tab:rq2-pruning} presents a comparison with alternative general VLM visual token pruning methods.
Specifically, we compare against Fast-V~\cite{chen2024fastv}, HiPrune~\cite{liu2025hiprunetrainingfreevisualtoken}, and Vision-Zip~\cite{yang2025visionzip}. 
Our \focusui\ preserves near-baseline accuracy at 30\% token retention (within 0.5/3.2/2.5 points on ScreenSpot-V2/Pro/OS-World-G), while general pruning severely degrades performance. Notably, our method is natively compatible with FlashAttention~\cite{dao2022flashattention} since it does not require any intermediate attention or activation information. 

\begin{table}[!t]
\centering
\small
\resizebox{\columnwidth}{!}{
\setlength{\tabcolsep}{1.5pt}
\begin{tabular}{lcccc}
\toprule
\textbf{Model} & \textbf{\%Ret.} & \textbf{SS-V2} & \textbf{SS-Pro} & \textbf{OSWorld-G}\\
\,\, + \textbf{Pruning Method (Venue)} & \textbf{ Ratio} & \textbf{Avg} & \textbf{Avg} & \textbf{Avg} \\
\midrule
Qwen2.5-VL-3B & $100\%$ & 81.5 & 26.1 & 27.3 \\
\,\, + Fast-V (ECCV'24)~\cite{chen2024fastv} & $30\%$ & 38.6 {\scriptsize (-52.7\%)} & 4.8 {\scriptsize (-81.6\%)} & 14.4 {\scriptsize (-47.4\%)} \\
\,\, + HiPrune (arXiv'25)~\cite{liu2025hiprunetrainingfreevisualtoken} & $30\%$ & 72.0 {\scriptsize (-11.7\%)} & 18.0 {\scriptsize (-30.8\%)} & 20.4 {\scriptsize (-25.3\%)} \\
\,\, + Vision-Zip (CVPR'25)~\cite{yang2025visionzip} & $30\%$ & 75.4 {\scriptsize (-7.5\%)} & 18.9 {\scriptsize (-27.4\%)} & 23.0 {\scriptsize (-15.6\%)} \\
\midrule
Jedi-3B & $100\%$ & 88.9 & 36.1 & 48.8 \\
\,\, + Fast-V (ECCV'24)~\cite{chen2024fastv} & $30\%$ & 51.0 {\scriptsize (-42.6\%)} & 14.1 {\scriptsize (-60.9\%)} & 23.9 {\scriptsize (-51.0\%)} \\
\,\, + HiPrune (arXiv'25)~\cite{liu2025hiprunetrainingfreevisualtoken} & $30\%$ & 80.9 {\scriptsize (-9.0\%)} & 26.2 {\scriptsize (-27.3\%)} & 40.4 {\scriptsize (-17.1\%)} \\
\,\, + Vision-Zip (CVPR'25)~\cite{yang2025visionzip} & $30\%$ & 82.8 {\scriptsize (-6.9\%)} & 28.8 {\scriptsize (-20.3\%)} & 41.5 {\scriptsize (-14.9\%)} \\
\midrule
\focusuithreebillion & $100\%$ & 91.5 & 43.8 & 53.4 \\
\,\, + Saliency Scorer w/ \textsc{PosPad} & $30\%$ & 91.0 {\scriptsize (-0.5\%)} & 40.6 {\scriptsize (-7.3\%)} & 51.8 {\scriptsize (-3.0\%)} \\
\bottomrule
\end{tabular}
}
\vspace{-0.5em}
\caption{Comparison to general visual token pruning methods.}
\vspace{-0.25em}
\label{tab:rq2-pruning}
\end{table}

\subsubsection{RQ3: Efficiency Analysis}
\label{sec:rq3-efficiency}
\begin{table}[!t]
\centering
\small
\resizebox{\columnwidth}{!}{  
\setlength{\tabcolsep}{2.5pt}
\begin{tabular}{lccccc}
\toprule
\multirow{2}{*}{\textbf{Model}} & \textbf{\%Ret.} & \textbf{\#Vis.} & \textbf{per Sample} & \textbf{Max GPU } & \textbf{SS-Pro} \\
& \textbf{ Ratio} & \textbf{Token} & \textbf{Time (sec)} & \textbf{Mem. (MB)} & \textbf{Acc} \\
\midrule
\rowcolor{gray!15}\multicolumn{6}{c}{\textit{Base Model: Qwen2.5-VL, $max\_pixel=6400*28*28=4816000$}} \\
\textcolor{gray!100}{\focusuisevenbillion} & \textcolor{gray!100}{$100\%$} & \textcolor{gray!100}{5319} & \textcolor{gray!100}{1.75 {\scriptsize ($1.00\times$)}} & \textcolor{gray!100}{20994 {\scriptsize ($1.00\times$)}} & 48.3 \\
\focusuisevenbillion & $70\%$  & 3989 & 1.67 {\scriptsize ($1.05\times$)} & 18334 {\scriptsize ($0.87\times$)} & 47.7 \\
\focusuisevenbillion & $50\%$  & 2659 & 1.49 {\scriptsize ($1.18\times$)} & 17944 {\scriptsize ($0.85\times$)} & 46.5 \\
\focusuisevenbillion & $30\%$  & 1329 & 1.22 {\scriptsize ($1.44\times$)} & 17392 {\scriptsize ($0.83\times$)} & 45.1 \\
\midrule
\rowcolor{gray!15}\multicolumn{6}{c}{\textit{Base Model: Qwen3-VL, $max\_pixel=6000*32*32=6144000$}} \\
\textcolor{gray!100}{\focusuitwobillionqwenthreevl} & \textcolor{gray!100}{$100\%$} & \textcolor{gray!100}{4627} & \textcolor{gray!100}{0.97 {\scriptsize ($1.00\times$)}} & \textcolor{gray!100}{6278 {\scriptsize ($1.00\times$)}} & 39.8 \\
\focusuitwobillionqwenthreevl & $70\%$  & 3470 & 0.90 {\scriptsize ($1.08\times$)} & 6142 {\scriptsize ($0.98\times$)} & 40.1 \\
\focusuitwobillionqwenthreevl & $50\%$  & 2313 & 0.85 {\scriptsize ($1.14\times$)} & 5680 {\scriptsize ($0.91\times$)} & 40.4 \\
\focusuitwobillionqwenthreevl & $30\%$  & 1156 & 0.71 {\scriptsize ($1.37\times$)} & 5170 {\scriptsize ($0.82\times$)} & 38.5 \\
\bottomrule
\end{tabular}
}
\vspace{-0.5em}
\caption{Efficiency analysis on ScreenSpot-Pro benchmark under different retention ratios and model backbones of \focusui. $^{*}$ The number of \pospadtoken\ tokens is not  included.}
\label{tab:rq2-efficiency}
\vspace{-1.5em}
\end{table}

In Tab.~\ref{tab:rq2-efficiency}, we evaluate \focusui\ with different Qwen2.5-VL and Qwen3-VL backbones to study efficiency gains and accuracy-efficiency trade-offs. Results show that reducing retention ratio from 100\% to 30\% yields up to 1.44$\times$ faster inference and about 17--18\% lower peak memory with only 3.2-point accuracy loss.

\begin{figure*}[!th]
    \centering
    \includegraphics[width=0.95\linewidth]{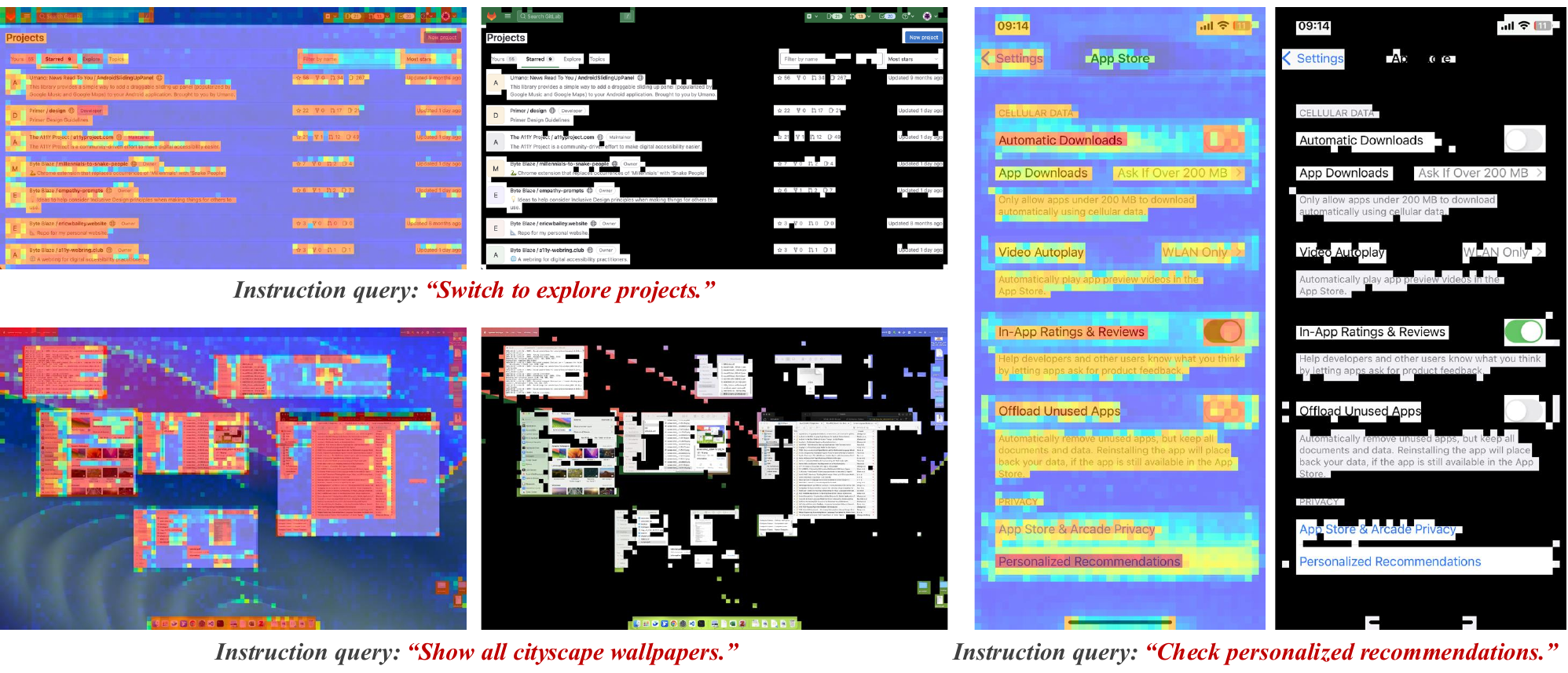}
    \vspace{-0.5em}
\caption{Qualitative visualization of predicted saliency heatmaps and retained patches under a \textbf{retention ratio} $r=\mathbf{30\%}$. Black regions denote dropped visual tokens that are not consumed by the LM during decoding. Examples are taken from the ScreenSpot-V2 and ScreenSpot-Pro benchmarks, spanning web, desktop, and mobile interfaces.}
\vspace{-0.75em}
\label{fig:scsp_vis_attend}
\end{figure*}

\subsubsection{RQ4: Qualitative Results}
\label{sec:rq4-qualitative}
Fig.~\ref{fig:scsp_vis_attend} shows qualitative examples of \focusui. The predicted heatmaps show that the model effectively selects the relevant visual tokens for the instruction while suppressing background regions.

\subsubsection{RQ5: Ablation Study}
\label{sec:rq5-ablation}

We highlight the effectiveness of our proposed components in Tab.~\ref{tab:rq5-ablation}. Models evaluated in experiments in Tab.~\ref{tab:rq5-ablation-selection} and~\ref{tab:rq5-ablation-ins2patch} use Qwen2.5-VL-3B as the base model and are trained with $30\%$ of the full data.

\begin{table}[!ht]
\centering
\begin{minipage}[t]{\columnwidth}
\resizebox{\columnwidth}{!}{
    \setlength{\tabcolsep}{3pt}
    \begin{tabular}{llccccc}
    \toprule
    \multirow{2}{*}{\textbf{Scoring Variant}}  & \textbf{Retain} & \textbf{\%Ret.} & \textbf{Reduce} & \textbf{Preserve Pos.} & \textbf{SS-Pro} \\
    & \textbf{Strategy} & \textbf{ Ratio} & \textbf{Token Len?} & \textbf{Continuity?} & \textbf{Acc} \\
    \midrule
    Baseline & (a) N/A & 100\% & \xmark & \cmark & 40.9 \\
    \midrule
    \multirow{3}{*}{CLIP Score~\cite{radford2021learning}} & (b) Direct drop & $50\%$ & \cmark & \xmark & 28.5 \\
    & (c) Full padding & $50\%$ & \xmark & \cmark & 38.7 \\
    & (d) \pospad & $50\%$ & \cmark & \cmark & 38.2 \\
    \midrule
    \multirow{3}{*}{Ins2Patch Score} & (b) Direct drop & $50\%$ & \cmark & \xmark & 29.2 \\
    & (c) Full padding & $50\%$ & \xmark & \cmark & 42.1 \\
    & (d) \pospad & $50\%$ & \cmark & \cmark & \textbf{42.3} \\
    \bottomrule
    \end{tabular}
}
\subcaption{Different visual token selection methods and positional continuity retention strategies.}
\vspace{-0.4em}
\label{tab:rq5-ablation-selection}
\end{minipage}
\begin{minipage}[t]{0.475\columnwidth}
\centering
\resizebox{\columnwidth}{!}{
    \setlength{\tabcolsep}{1pt}
    \begin{tabular}{lc}
    \toprule
    \textbf{Variant} & \textbf{SS-Pro Acc} \\
    \midrule
    w/ UI-Graph Labeling only & 41.1  \\
    w/ BBox-based Labeling only & 39.8 \\
    \midrule
    Full \focusui & \textbf{42.3} \\
    \bottomrule
    \end{tabular}
}
\subcaption{Ins2Patch score ablation with a reduction rate of 50\%.}
\label{tab:rq5-ablation-ins2patch}
\end{minipage}
\hfill
\begin{minipage}[t]{0.51\columnwidth}
\centering
\resizebox{\columnwidth}{!}{
    \setlength{\tabcolsep}{3pt}
    \begin{tabular}{lcccc}
    \toprule
    \textbf{\%Ret.} & \textbf{\#Vis.} & \textbf{\#\pospad} & \textbf{\#Total} & \textbf{SS-Pro} \\
    \textbf{Ratio} & \textbf{Tokens} & \textbf{Tokens} & \textbf{Tokens} & \textbf{Acc} \\
    \midrule
    $100\%$ & 6019 & 0 & 6140 & 43.8 \\
    $75\%$ & 4514 & 435 & 5070 & 43.3 \\
    $50\%$ & 3009 & 433 & 3563 & 42.3 \\
    $25\%$ & 1504 & 315 & 1941 & 40.6 \\
    $10\%$ & 601 & 193 & 915 & 36.6 \\
    \bottomrule
    \end{tabular}
}
\subcaption{Different retention ratios and numbers of tokens.}
\label{tab:rq5-ablation-ratios}
\end{minipage}
\vspace{-0.5em}
\caption{Ablation of key components of \focusui.}
\label{tab:rq5-ablation}
\vspace{-1em}
\end{table}

\noindent\textbf{Visual Token Selection.}\label{sec:rq5-ablation-selection}
We compare with the variants illustrated in Fig.~\ref{fig:pospad}:
(a) \emph{Original visual sequence}.
(b) \emph{Direct Drop}. 
(c) \emph{Full Padding} which preserves continuity by inserting \pospadtoken\ at every dropped position. 
We also test zero-shot CLIP~\cite{radford2021learning} as the scoring strategy.
The performance of these variants is shown in Tab.~\ref{tab:rq5-ablation-selection}.

\noindent\textbf{Instruction-to-Patch Saliency Score Supervision.}\label{sec:rq5-ablation-ins2patch}
Results in Tab.~\ref{tab:rq5-ablation-ins2patch} indicate that removing either the UI-graph prior or the bounding-box overlap score degrades accuracy relative to the fused supervision of \focusui.

\noindent\textbf{Retention Ratio.}\label{sec:rq5-ablation-retention}
Tab.~\ref{tab:rq5-ablation-ratios} suggests a smooth accuracy-retention trade-off of our \focusui: 100\% matches the dense baseline, 50\% still retains most performance, and further aggressive settings incur larger accuracy drops.

\vspace{-0.25em}
\section{Related Work}
\label{sec:related}

\paragraph{VLM-Powered GUI Agents}
Recent advances in VLMs have accelerated progress on GUI agents that perceive, plan, and act in graphical interfaces~\cite{wang2024qwen2vl,bai2025qwen2_5vl,anthropic2025claude,openai2025operator,google2024gemini}.
Starting from text-dependent GUI agents~\cite{zhou2023webarena, gao2023assistgui}, it progressively transitions to purely visual solutions for task planning, element grounding, and interface control~\cite{cheng2024seeclick,lin2024showui,qin2025ui,xie2025scaling} that fully utilizes VLMs' capability.

\paragraph{UI Visual Grounding}
Given a screenshot and a natural language instruction, UI grounding locates the target region for interaction on the screen.
With more advanced model design~\cite{ wu2024atlas,cheng2024seeclick,lin2024showui,ariaui,qin2025ui,zhang2025tongui,tang2025think,wu2025gui,chen2025v2pbackgroundsuppressioncenter} and data scaling~\cite{chen2024guicourse,xu2024aguvis,gou2024navigating,xie2025scaling}, the performance of UI grounding improves rapidly in recent times.

\paragraph{Visual Token Reduction}
Compared to information-dense text, visual tokens often exhibit substantial redundancy, motivating token reduction to lower computation cost~\cite{xing2025pyramiddropacceleratinglargevisionlanguage,bolya2022tome,chen2024fastv, zhang2024fastervlm}.
Recent work further explores training-free pruning based on token importance and redundancy~\cite{zhang2025sparsevlm, liu2025hiprunetrainingfreevisualtoken} or implement encoder-side compression~\cite{yang2025visionzip,zhang2025vscanrethinkingvisualtoken}.


\vspace{-0.25em}
\section{Conclusion}
\label{sec:conclusion}
In this paper, we introduced \focusui, a query-guided framework for efficient UI grounding that selects \emph{instruction-relevant} visual tokens while preserving positional continuity. Integrated with state-of-the-art VLMs, \focusui\ achieves strong accuracy-efficiency trade-offs across four UI grounding benchmarks.
\vspace{-0.25em}
\paragraph{Limitations and Future Work.} \focusui\ primarily gains efficiency from spatial visual token reduction. Future work may consider the temporal dimension, as UI interactions typically involve multi-round and sequential actions.

\appendix
\setcounter{section}{0}

\section{Implementation Details}
\label{sec:impl_details}

\subsection{Training Data}

\paragraph{Raw Dataset} Our training set compiles several public high-quality GUI datasets, following GUI-Actor. To ensure fair evaluation, samples from Wave-UI that overlap with the test sets of downstream tasks are excluded.

\paragraph{Refining Annotation Quality}
We apply OmniParser V2~\cite{lu2024omniparser} to filter samples whose IoU between ground-truth and OmniParser detected boxes is 
below 0.3. This results in a reduction of 22.9\% in the number of elements. The final training statistics are in Tab.~\ref{tab:supp_training-data}.

\begin{table}[h]
\resizebox{\columnwidth}{!}{  
\centering
\setlength{\tabcolsep}{4pt}
\begin{tabular}{lcccc}
\toprule
\textbf{Dataset} & \textbf{\#Screenshots} & \textbf{\#Elements} & \textbf{Platform} \\
\midrule
UGround~\cite{gou2024navigating}  & 775K & 8M   & Web \\
GUI-Env~\cite{wu2025gui} & 70K  & 262K & Web \\
GUI-Act~\cite{wu2025gui} & 13K  & 42K  & Web \\
AndroidControl~\cite{li2024effects} & 47K  & 47K  & Android \\
AMEX~\cite{chai2024amex} & 100K & 1.2M & Android \\
Wave-UI & 7K   & 50K  & Hybrid \\
\midrule
\textbf{Total (Raw Dataset)} & 1012K & 9.6M & -- \\
\textbf{Total (After Filtering)} & \textbf{976K} & \textbf{7.4M} & -- \\
\bottomrule
\end{tabular}
}
\caption{Statistics of training datasets used for \focusui.}
\label{tab:supp_training-data}
\end{table}

\subsection{Training Details}
We train \focusui\ on $8\times$ NVIDIA H200 GPUs using bfloat16 precision, DeepSpeed ZeRO-2~\cite{10.1145/3394486.3406703}, and FlashAttention-2~\cite{dao2022flashattention}. The effective batch size per GPU is set to 32 (with gradient accumulation of 4), and the \texttt{max\_pixels} is set to $5720064$, matching GUI-Actor. Training proceeds in two stages:

\noindent \textbf{Stage 1: Saliency Scorer Pre-training.} We pretrain the randomly initialized Query-Guided Saliency Scorer for 1 epoch with a learning rate of $1e-4$. This takes about 12 hours for both 3B and 7B models.

\noindent \textbf{Stage 2: Full Model Fine-tuning.} We fine-tune all parameters for 1 epoch with a learning rate of $5e-6$. This takes about 36 hours for the 3B model and about 48 hours for the 7B model.

\noindent \textbf{Hyperparameter Details.} During training, we construct per-patch saliency score supervision with $\tau=2$ and $\lambda=0.8$. Patch size $p$ is set to 14 and 16 for Qwen2.5-VL and Qwen3-VL based models, respectively. The visual token retention ratio $r$ is uniformly sampled from $(0.1, 1.0)$ for each training sample. 

\begin{figure}[!t]
  \centering
  \begin{subfigure}[t]{\linewidth}
      \centering
      \includegraphics[width=\linewidth]{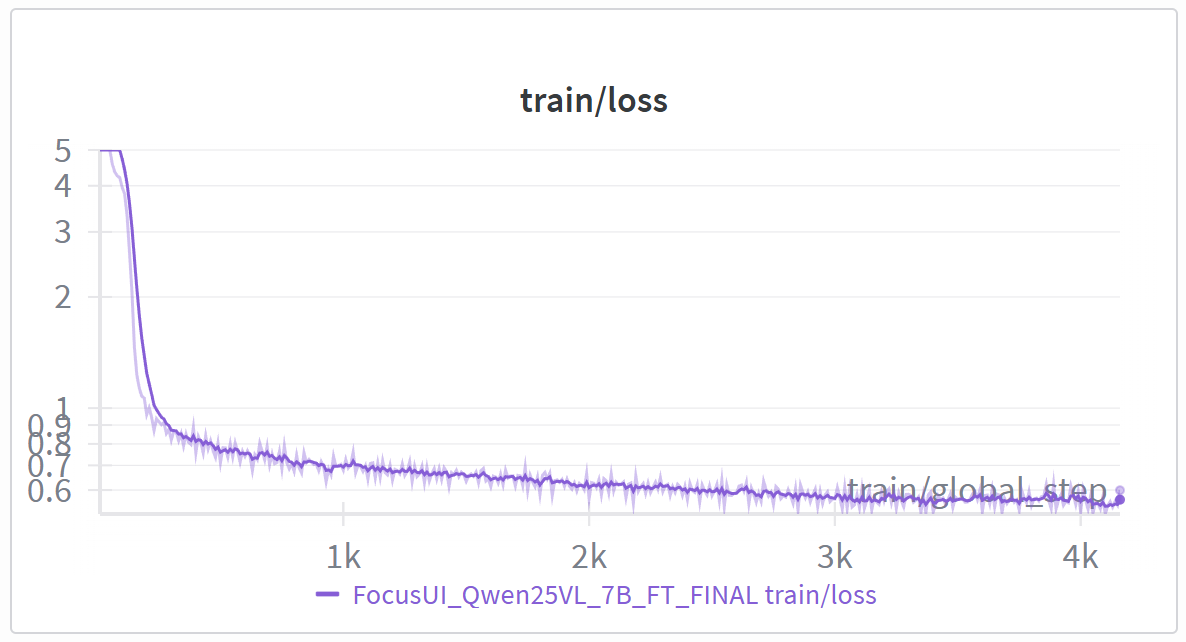}
      \caption{\textbf{Total Loss} curve during training.}
      \label{fig:supp_wandb_loss}
  \end{subfigure}
  \vfill
  \begin{subfigure}[t]{\linewidth}
      \centering
      \includegraphics[width=\linewidth]{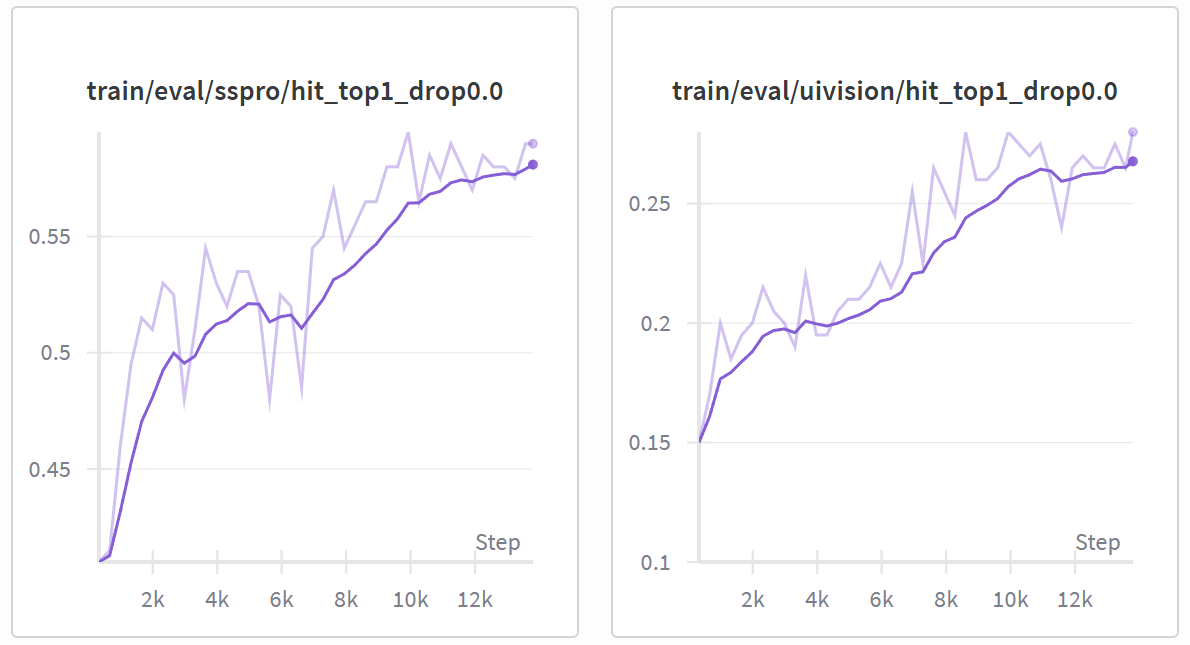}
      \caption{\textbf{Evaluation:} ScreenSpot-Pro and UI-Vision with \textbf{retention ratio = 100\%}.}
      \label{fig:supp_wandb_eval}
  \end{subfigure}
  \vfill
  \begin{subfigure}[t]{\linewidth}
      \centering
      \includegraphics[width=\linewidth]{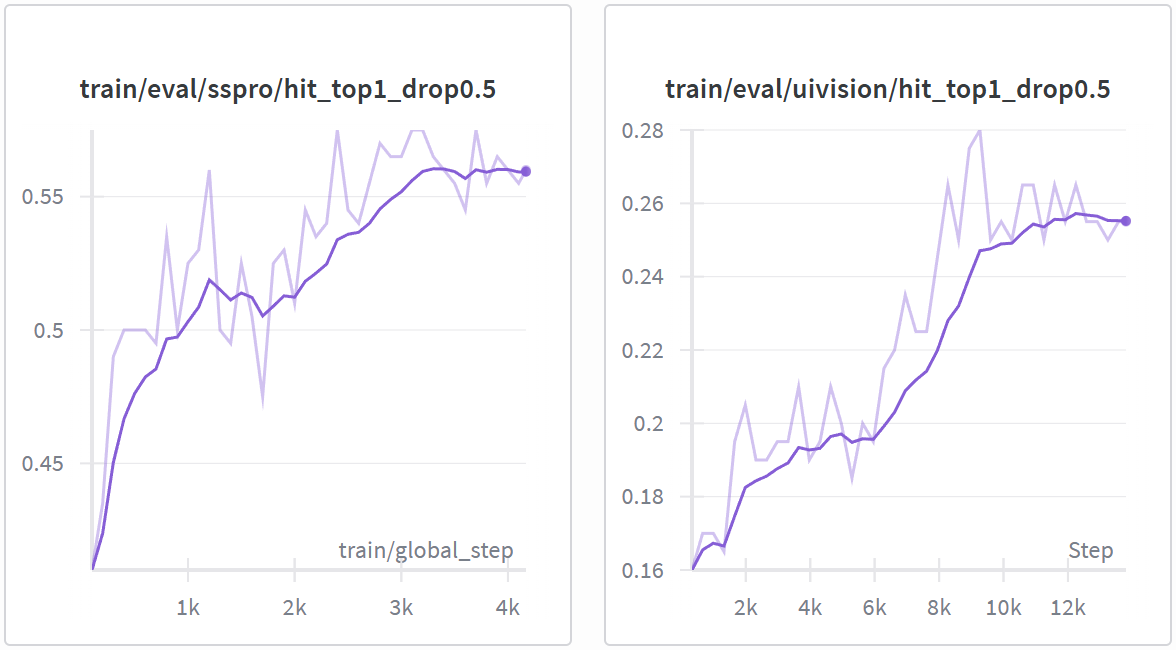}
      \caption{\textbf{Evaluation:} ScreenSpot-Pro and UI-Vision with \textbf{retention ratio = 50\%}.}
      \label{fig:supp_wandb_eval_retain02}
  \end{subfigure}
  \caption{WandB loss and evaluation results of \focusuisevenbillion.}
  \label{fig:supp_wandb}
  \vspace{-1em}
\end{figure}

To enable reproducibility, we use the final checkpoint for all obtained \focusui\ models. We also provide the full Weights \& Biases (WandB) logs 
for all trained models. The training loss and evaluation curves during the training of \focusuisevenbillion\ are shown in Fig.~\ref
{fig:supp_wandb}.

\subsection{UI Grounding Benchmarks}  
We evaluate on four public benchmarks containing screenshots paired with instructions: \textbf{ScreenSpot-V2}~\cite{wu2024atlas}, \textbf{ScreenSpot-Pro}~\cite{li2025screenspot}, \textbf{OS-World-G}~\cite{xie2025scaling}, and \textbf{UI-Vision}~\cite
{nayak2025uivisiondesktopcentricguibenchmark}. The statistics of these benchmarks are shown in Tab.~\ref{tab:testing-benchmarks}.

\noindent\textbf{ScreenSpot-V2~\cite{wu2024atlas}.} A refined version of ScreenSpot~\cite{cheng2024seeclick} with 1,272 samples across mobile, desktop, and web environments.

\noindent\textbf{ScreenSpot-Pro~\cite{li2025screenspot}.} This benchmark contains 1,581 samples from 23 professional applications, targeting high-resolution interfaces and complex layouts to test generalization.

\noindent\textbf{OS-World-G~\cite{xie2025scaling}.} Sampled from OSWorld~\cite{xie2024osworld}, this benchmark includes 564 samples categorized by task type (text matching, element recognition, layout understanding, fine-grained manipulation, and refusal).

\noindent\textbf{UI-Vision~\cite{nayak2025uivisiondesktopcentricguibenchmark}.} A desktop-centric benchmark with 5,790 samples from 83 applications, evaluating element grounding, layout grounding, and action prediction.

\begin{table}[h]
\resizebox{\columnwidth}{!}{
  \centering
  \setlength{\tabcolsep}{3pt}
  \begin{tabular}{lcccc}
    \toprule
    \textbf{Benchmark} & \textbf{\#Samples} & \textbf{Avg Res.} & \textbf{Max Res.} & \textbf{Platform} \\
    \midrule
    ScreenSpot-V2 & 1272 & 1725$\times$1657 & 2880$\times$1800 & Hybrid \\
    ScreenSpot-Pro & 1581 & 3267$\times$1727 & 6016$\times$3384 & Desktop \\
    OS-World-G & 564 & 1696$\times$955 & 1920$\times$1080 & Desktop \\
    UI-Vision & 5790 & 1851$\times$1034 & 3360$\times$2036 & Desktop \\
    \bottomrule
  \end{tabular}
}
\caption{Overview of the evaluation benchmarks used in this work.}
\label{tab:testing-benchmarks}
\vspace{-1em}
\end{table}

\section{Discussion}
\label{sec:discussion_appendix}

\subsection{Visual Redundancy Analysis}
Tab.~\ref{tab:supp_vis-redundancy} provides the token statistics of \textbf{Study 1} (shown in Fig.~1(b) in the main paper). Using the default model settings on the ScreenSpot-Pro evaluation, we find that visual tokens occupy at least 84.3\% of the sequence across the studied benchmarks, confirming significant visual redundancy in UI grounding tasks.

\begin{table}[h]
  \resizebox{\columnwidth}{!}{
    \centering
    \setlength{\tabcolsep}{3pt}
    \begin{tabular}{llcccc}
      \toprule
      \multirow{2}{*}{\textbf{Benchmark}} &  \multirow{2}{*}{\textbf{Model}} & \textbf{\#Sys.} & \textbf{\#Vis.} & \textbf{\#Inst.} & \textbf{Vis. Token} \\
      & & \textbf{Tokens} & \textbf{Tokens} & \textbf{Tokens} & \textbf{\%} \\
      \midrule
      \multirow{2}{*}{\textbf{ScreenSpot-V2}} & Jedi-1080p & 397 & 2348 & 4.5 & 85.4\% \\
        & GUI-Actor & 90 & 3506 & 4.5 & 97.1\% \\
      \midrule
      \multirow{2}{*}{\textbf{ScreenSpot-Pro}} & Jedi-1080p & 397 & 2629 & 5.2 & 86.7\% \\
        & GUI-Actor & 90 & 5801 & 5.2 & 98.1\% \\
      \midrule
      \multirow{2}{*}{\textbf{OS-World-G}} & Jedi-1080p & 397 & 2244 & 21.3 & 84.3\% \\
        & GUI-Actor & 90 & 2244 & 21.3 & 95.3\% \\
      \midrule
      \multirow{2}{*}{\textbf{UI-Vision}} & Jedi-1080p & 397 & 2249 & 9.9 & 84.7\% \\
        & GUI-Actor & 90 & 2566 & 9.9 & 96.3\% \\
      \bottomrule
    \end{tabular}
  }
  \caption{Token statistics of \textbf{Study 1} shown in Fig.~1(b) in the main paper.}
  \label{tab:supp_vis-redundancy}\vspace{-1em}
\end{table}

\subsection{Position Sensitivity Analysis}

Tab.~\ref{tab:supp_pruning} shows the detailed results of \textbf{Study 2} (shown in FIg.~1(c) in the main paper), comparing \focusui\ with UI grounding models integrated with advanced visual token pruning methods.

\begin{table}[ht]
  \centering
  \small
  \resizebox{\columnwidth}{!}{
  \setlength{\tabcolsep}{6pt}
  \begin{tabular}{ll@{\hspace{6pt}}ccc}
  \toprule
  \textbf{\%Ret.} & \textbf{Model} & \textbf{SS-V2} & \textbf{SS-Pro} & \textbf{OSW-G}\\
  \textbf{ Ratio} & \,\, + \textbf{Pruning Method} & \textbf{Avg} & \textbf{Avg} & \textbf{Avg} \\
  \midrule
  $100\%$ & Qwen2.5-VL-3B & 81.5 & 26.1 & 27.3 \\
  \cmidrule(lr){1-2} \multirow{3}{*}{$50\%$} &\,\, + Fast-V  & 43.5 & 13.9 &  14.3 \\
  &\,\, + HiPrune & 80.4 & 20.3 & 26.2  \\
  &\,\, + Vision-Zip & 81.0 & 21.0 &  27.1 \\
  \cmidrule(lr){1-2} \multirow{3}{*}{$30\%$} &\,\, + Fast-V  & 38.6 & 4.8 & 14.4  \\
  &\,\, + HiPrune & 72.0 & 18.0  & 20.4  \\
  &\,\, + Vision-Zip & 75.4 & 18.9 & 23.0 \\
  \midrule
  $100\%$ & Jedi-3B & 88.9 & 36.1 & 48.8 \\
  \cmidrule(lr){1-2} \multirow{3}{*}{$50\%$} &\,\, + Fast-V & 50.3 & 20.4 &  25.3 \\
  &\,\, + HiPrune & 88.3 & 32.8 &  46.4  \\
  &\,\, + Vision-Zip & 88.1 & 32.9 &  46.6   \\
  \cmidrule(lr){1-2} \multirow{3}{*}{$30\%$} &\,\, + Fast-V & 51.0 & 14.1 & 23.9 \\
  &\,\, + HiPrune & 80.9 & 26.2 & 40.4 \\
  &\,\, + Vision-Zip & 82.8 & 28.8 & 41.5 \\
  \midrule
  $100\%$ &  \focusuithreebillion & 91.5 & 43.8 & 53.4 \\
  \cmidrule(lr){1-2} $50\%$ & \,\, + Full Settings & 91.4 & 42.3 & 54.6 \\
  $30\%$ & \,\, + Full Settings & 91.0 & 40.6 & 51.8 \\
  \bottomrule
  \end{tabular}
  }
  \caption{Detailed comparison with general visual token pruning methods for \textbf{Study 2} shown in Fig.~1(c) in the main paper.}
  \label{tab:supp_pruning}
\end{table}

\section{More Experimental Results}

\subsection{Effective Visual Selection: Patch Recall@K\%}

We verify the effectiveness of our visual token selection using \textbf{Patch Recall@K\%}, defined as the fraction of ground-truth (GT) regions captured within the top K\% of saliency-ranked patches (\ie, the top-K visual tokens):
$$
\text{Patch Recall@}K\% \;=\;
\frac{\left|\,\text{GT positive regions in top }K\%\,\right|}
     {\left|\,\text{Total GT positive regions}\,\right|}
$$
where ground-truth positive regions correspond to the area of UI elements paired with the given instruction. We evaluate $K \in \{5\%, 10\%, 25\%, 50\%\}$. Additionally, we report the \textbf{Full Coverage Budget}, the percentage of visual tokens needed to fully cover the ground-truth elements. Results are shown in Tab.~\ref{tab:patchrecall}.

\begin{table}[h]
\resizebox{\columnwidth}{!}{
\centering
  \setlength{\tabcolsep}{2pt}
  \begin{tabular}{l|ccccc|c}
    \toprule
    \multirow{2}{*}{\textbf{Model}} & \multicolumn{5}{c|}{\textbf{Patch Recall@K\% $\uparrow$}} & \textbf{Full Coverage} \\ 
    & \textbf{@5\%} & \textbf{@10\%} & \textbf{@25\%} & \textbf{@50\%} & \textbf{Avg} & \textbf{Budget $\downarrow$}\\ 
    \midrule 
    \rowcolor{gray!15}\multicolumn{7}{c}{\textit{Zero-shot Baselines}} \\
    Random & 0.05 & 0.11 & 0.26 & 0.51 & 0.23 & 0.85 \\
    CLIP & 0.12 & 0.21 & 0.41 & 0.65 & 0.35 & 0.61 \\
    \midrule 
    \rowcolor{gray!15}\multicolumn{7}{c}{\textit{Our Query-Guided Saliency Scorer}} \\
    \focusuithreebillion & 0.39 & 0.56 & 0.83 & 0.96 & 0.69 & 0.25 \\
    \focusuisevenbillion & 0.43 & 0.60 & 0.84 & 0.97 & 0.71 & 0.24 \\
    \bottomrule
  \end{tabular}
}
  \caption{\textbf{Patch Recall@K\%} and \textbf{Full Coverage Budget} performance comparison on the ScreenSpot-Pro benchmark.}
  \label{tab:patchrecall}\vspace{-1em}
\end{table}

\subsection{Analysis of \pospad}

To better understand the effect of preserving the original spatial layout, we further analyze the placement of \pospad\ within contiguous sequences of dropped visual tokens, comparing three variants: (i) \emph{sequence-first}, (ii) \emph{sequence-middle}, and (iii) \emph{sequence-end} (our proposed \pospad).

\begin{table}[th]
  \resizebox{\columnwidth}{!}{
  \centering
  \setlength{\tabcolsep}{4pt}
  \begin{tabular}{lccccc}
    \toprule
    \multirow{2}{*}{\textbf{Sequence Type}} & \multicolumn{4}{c}{\textbf{SS-Pro Avg.}} \\
    \cmidrule(lr){2-5} & \textbf{$r=100\%$} & \textbf{$r=75\%$} & \textbf{$r=50\%$} & \textbf{$r=25\%$} \\
    \midrule 
    sequence-first  & 42.0 & 41.8 & 39.8 & 36.8 \\ 
    sequence-middle  & 41.2 & 41.1 & 38.7 & 33.5 \\ 
    sequence-end (\pospad)  & 42.3 & 42.1 & 40.4 & 37.7 \\ 
    \bottomrule
  \end{tabular}
  }
  \caption{Different placement of the \pospad\ token.}
  \label{tab:supp_pospad_ablation}\vspace{-1em}
\end{table}

\cref{tab:supp_pospad_ablation} reports a study that varies the visual token retention ratio $r$ and the location of the \pospad\ token.
Across all retention ratios, placing \pospad\ at the end of each dropped sequence achieves the best performance, especially under 
low retention ratios. The empirical results confirm our intuition: placing \pospad\ at the end of the sequence is more compatible with the raster-scan ordering used by the vision encoder and M-RoPE. In contrast, placing \pospad\ at the beginning or in the middle of the 
sequence pulls the whole region toward earlier positions, making it harder for the LM decoder to align with the original spatial 
structure.

\subsection{Detailed Performance vs. Retention Ratio}

\begin{figure}[ht]
  \centering
  \begin{subfigure}[t]{\linewidth}
      \centering
      \includegraphics[width=\linewidth]{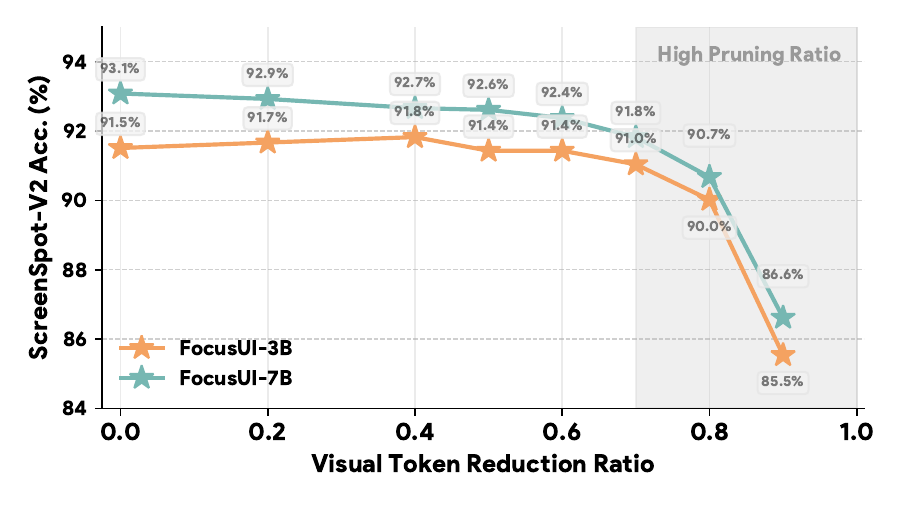}
      \caption{Performance vs. reduction ratio on ScreenSpot-V2.}
      \label{fig:supp_2b}
  \end{subfigure}
  \vfill
  \begin{subfigure}[t]{\linewidth}
      \centering
      \includegraphics[width=\linewidth]{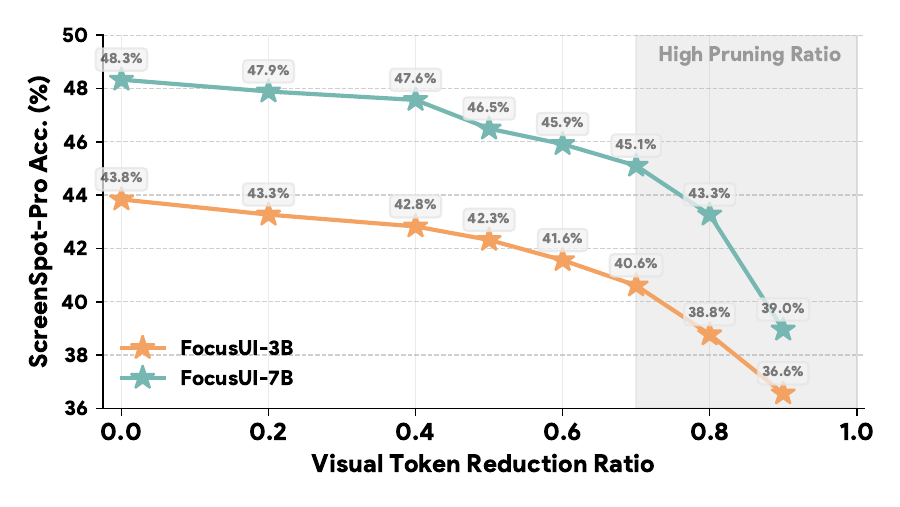}
      \caption{Performance vs. reduction ratio on ScreenSpot-Pro.}
      \label{fig:supp_2a}
  \end{subfigure}
  \caption{UI grounding accuracy under different token reduction ratios.}
  \label{fig:supp_performance_vs_retention}
  \vspace{-1em}
\end{figure}

Fig.~\ref{fig:supp_performance_vs_retention} presents the detailed performance of \focusui\ on ScreenSpot-V2 and ScreenSpot-Pro across varying reduction ratios ($1-\text{retention ratio}~r$). 
The results demonstrate that \focusui\ maintains high UI grounding accuracy even with significant visual token reduction.

\subsection{Qualitative Examples}

\begin{figure*}[!t]
  \centering
  \begin{subfigure}[t]{\linewidth}
    \centering
    \includegraphics[width=1.\linewidth]{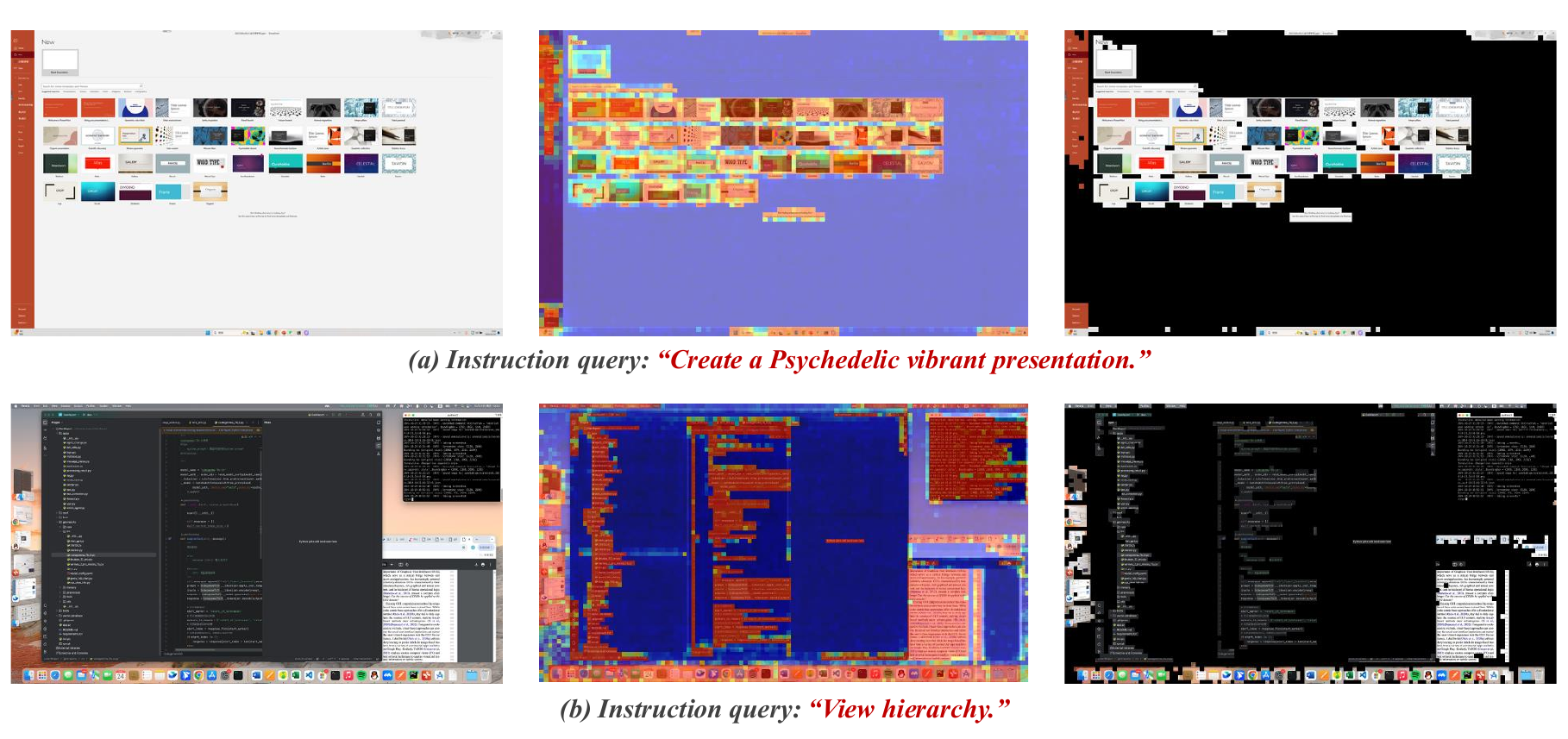}
    \vspace{-2em}
  \end{subfigure}
  \begin{subfigure}[t]{\linewidth}
  \centering
  \includegraphics[width=1.\linewidth]{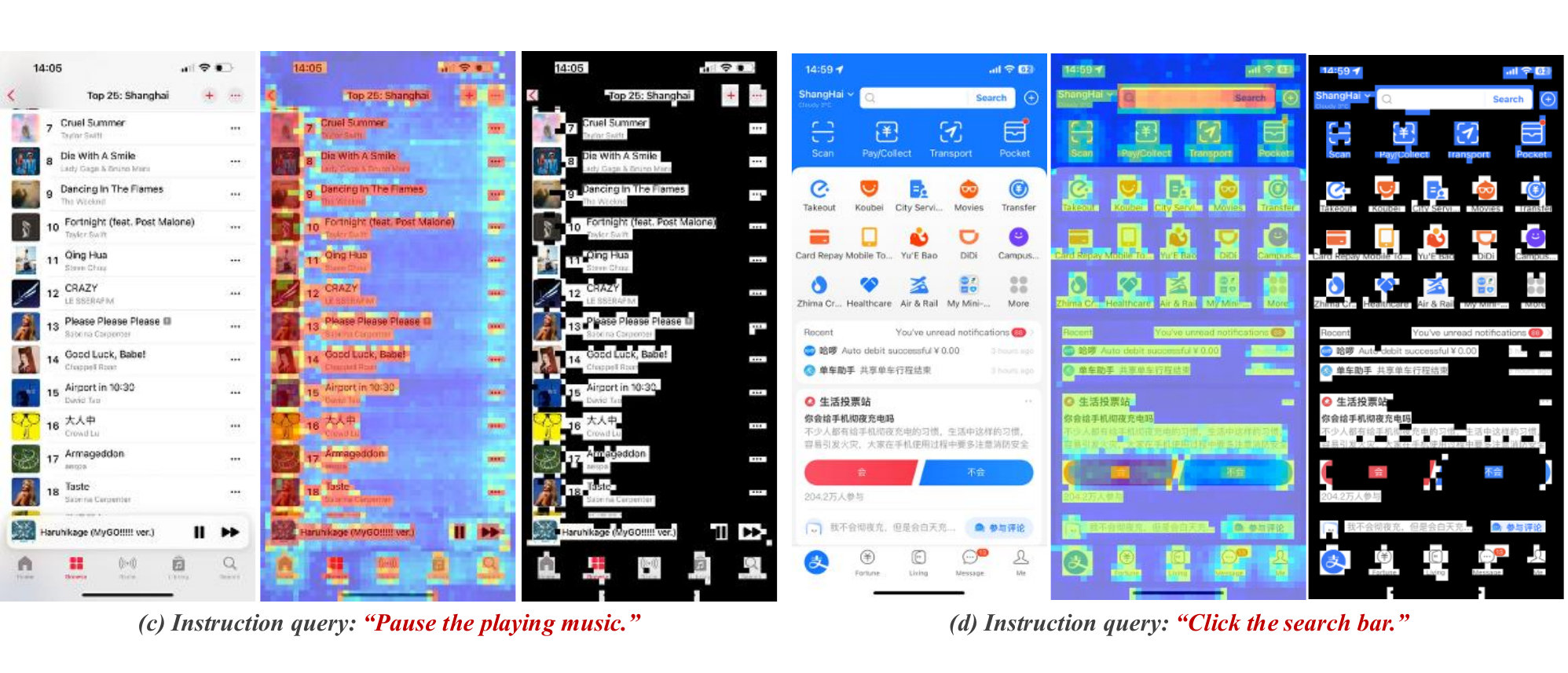}
  \end{subfigure}
  \vspace{-2em}
  \caption{Qualitative examples of predicted per-patch saliency. \textbf{Left:} original screenshot; \textbf{Middle:} predicted saliency map; and \textbf{Right:} visual token selection results with $r=30\%$.}
  \label{fig:supp_qual}
  \vspace{-1em}
\end{figure*}

Fig.~\ref{fig:supp_qual} visualizes saliency maps on ScreenSpot-V2 and ScreenSpot-Pro, showing that our Query-Guided Saliency Scorer effectively highlights instruction-relevant regions while suppressing the background. We find that for straightforward tasks (Fig.~\ref{fig:supp_qual} (a, d)), saliency scores peak significantly at the ground-truth locations. In more complex scenarios (Fig.~\ref{fig:supp_qual} (b, c)), while the scores may be less concentrated, the model still successfully distinguishes potential targets from irrelevant background elements.

\section{Prompt Templates}

\paragraph{\focusui\ (with Qwen2.5-VL base model)}
Below is the system prompt for \focusuithreebillion\ and \focusuisevenbillion.

\lstset{style=promptstyle}

\begin{lstlisting}
You are a GUI agent. Given a screenshot of the current GUI and a human instruction, your task is to locate the screen element that corresponds to the instruction. You should output a PyAutoGUI action that performs a click on the correct position. To indicate the click location, we will use some special tokens, which is used to refer to a visual patch later. For example, you can output: pyautogui.click(<your_special_token_here>).
\end{lstlisting}

\paragraph{\focusui\ (with Qwen3-VL base model)}

Below is the system prompt for \focusuitwobillionqwenthreevl.

\begin{lstlisting}
You are a GUI agent. Your task is to locate the screen element that corresponds to the instruction. You should not call any external tools. Output only the coordinate of one point in your response. Format: (x, y)

\end{lstlisting}

\paragraph{Qwen2.5-VL}

Below is the system prompt for evaluating Qwen2.5-VL models.

\begin{lstlisting}
You are a GUI agent. Your task is to locate the screen element that corresponds to the instruction. You should not call any external tools. Output only the coordinate of one point in your response. Format: (x, y)

\end{lstlisting}

\paragraph{Jedi and Qwen3-VL}

Below is the system prompt for evaluating Jedi-3B, Jedi-7B, and Qwen3-VL models.

\begin{lstlisting}
# Tools
You may call one or more functions to assist with the user query.

You are provided with function signatures within <tools></tools> XML tags:
<tools>
{"type": "function", "function": {"name": "computer_use", "description": "Use a mouse to interact with a computer.\n* The screen's resolution is {screen_width}x{screen_height}.\n* Make sure to click any buttons, links, icons, etc with the cursor tip in the center of the element. Don't click boxes on their edges unless asked.\n* you can only use the left_click and mouse_move action to interact with the computer. if you can't find the element, you should terminate the task and report the failure.", "parameters": {"properties": {"action": {"description": "The action to perform. The available actions are:\n* `mouse_move`: Move the cursor to a specified (x, y) pixel coordinate on the screen.\n* `left_click`: Click the left mouse button with coordinate (x, y).\n* `terminate`: Terminate the current task and report its completion status.", "enum": ["mouse_move", "left_click"], "type": "string"}, "coordinate": {"description": "(x, y): The x (pixels from the left edge) and y (pixels from the top edge) coordinates to move the mouse to. Required only by `action=mouse_move` and `action=left_click`.", "type": "array"}, "status": {"description": "The status of the task. Required only by `action=terminate`.", "type": "string", "enum": ["success", "failure"]}}, "required": ["action"], "type": "object"}}}
</tools>

For each function call, return a json object with function name and arguments within <tool_call></tool_call> XML tags:
<tool_call>
{"name": <function-name>, "arguments": <args-json-object>}
</tool_call>
\end{lstlisting}

{
\small
\bibliographystyle{ieeenat_fullname}
\bibliography{main}
}

\end{document}